\documentclass[10pt,journal,compsoc,twoside]{IEEEtran}
\IEEEoverridecommandlockouts
\usepackage{cite}
\usepackage{amsmath,amssymb,amsfonts}
\usepackage{algorithmic}
\usepackage{graphicx}
\usepackage{epstopdf}
\usepackage{textcomp}
\usepackage{bm}
\usepackage{stfloats}
\usepackage{amsthm}
\usepackage{url}
\usepackage[dvipsnames]{xcolor}
\usepackage[ruled,vlined]{algorithm2e}
\newtheorem{definition}{Definition}
\newtheorem{conjecture}{Conjecture}
\definecolor{mycolor}{RGB}{0,66,145}
\def\BibTeX{{\rm B\kern-.05em{\sc i\kern-.025em b}\kern-.08em
    T\kern-.1667em\lower.7ex\hbox{E}\kern-.125emX}}

\begin{document}

\title{Can We Enhance the Quality of Mobile Crowdsensing Data Without Ground Truth?}

\author{Jiajie~Li,
	Bo~Gu, ~\IEEEmembership{Member,~IEEE,}
	Shimin~Gong, ~\IEEEmembership{Member,~IEEE,}
	Zhou~Su, ~\IEEEmembership{Senior~Member,~IEEE,}
	\\and Mohsen Guizani, ~\IEEEmembership{Fellow,~IEEE}

\IEEEcompsocitemizethanks{
	\IEEEcompsocthanksitem This work was supported in part by the National Science Foundation of China (NSFC) under Grant U20A20175. \protect(Corresponding author: Bo Gu.)
	
	\IEEEcompsocthanksitem Jiajie Li, Bo Gu, and Shimin Gong are with the School of Intelligent Systems Engineering, Sun Yat-sen University, Shenzhen 518107, China, and with the Guangdong Provincial Key Laboratory of Fire Science and Intelligent Emergency Technology, Guangzhou 510006, China. \protect
		E-mail: lijj228@mail2.sysu.edu.cn, gubo@mail.sysu.edu.cn, and gong0012@e.ntu.edu.sg.
	
	\IEEEcompsocthanksitem Zhou Su is with the School of Cyber Science and Engineering, Xi'an Jiaotong University, Xi'an 710049, China. \protect
		E-mail: zhousu@ieee.org.
	
	\IEEEcompsocthanksitem Mohsen Guizani is with the Computer Science and Engineering Department, Qatar University, Doha 2713, Qatar. \protect E-mail: mguizani@ieee.org.
	}
}

\IEEEtitleabstractindextext{
\begin{abstract}
Mobile crowdsensing (MCS) has emerged as a prominent trend across various domains. However, ensuring the quality of the sensing data submitted by mobile users (MUs) remains a complex and challenging problem. To address this challenge, an advanced method is needed to detect low-quality sensing data and identify malicious MUs that may disrupt the normal operations of an MCS system. Therefore, this article proposes a prediction- and reputation-based truth discovery (PRBTD) framework, which can separate low-quality data from high-quality data in sensing tasks. First, we apply a correlation-focused spatio-temporal Transformer network that learns from the historical sensing data and predicts the ground truth of the data submitted by MUs. However, due to the noise in historical data for training and the bursty values within sensing data, the prediction results can be inaccurate. To address this issue, we use the implications among the sensing data, which are learned from the prediction results but are stable and less affected by inaccurate predictions, to evaluate the quality of the data. Finally, we design a reputation-based truth discovery (TD) module for identifying low-quality data with their implications. Given the sensing data submitted by MUs, PRBTD can eliminate the data with heavy noise and identify malicious MUs with high accuracy. Extensive experimental results demonstrate that the PRBTD method outperforms existing methods in terms of identification accuracy and data quality enhancement.
\end{abstract}

}
\maketitle
\IEEEpeerreviewmaketitle

\section{Introduction}
\label{sec:introduction}
\IEEEPARstart{M}{obile} crowdsensing (MCS) has drawn considerable attention in recent years. As a novel data acquisition paradigm, MCS leverages embedded sensors in mobile devices (e.g., smartphones and cameras) to gather large amounts of data for various applications \cite{survey}. Typically, an MCS system comprises three key types of components: a sensing platform, task initiators (TIs), and mobile users (MUs). Specifically, the TIs delegate sensing tasks to MUs through the sensing platform. The sensing platform gathers and sends the data submitted by the MUs to the TIs, subsequently assigning payments to the MUs. Compared with traditional sensing methods, MCS has advantages such as deployment flexibility, cost-effectiveness, extensive coverage, and scalability. Consequently, numerous applications based on MCS (e.g., ParkNet\cite{parknet}, Nextdoor\cite{nextdoor}, and Google Crowdsource\cite{googlecrowdsource}) have emerged.

However, the sensing data submitted by MUs may be unreliable because of sensor damage and the presence of malicious MUs \cite{decentralized,pp-assign,ppta,dqnmcs}. These low-quality sensing data (e.g., erroneous or outdated data) can mislead and negatively impact applications and systems based on sensing data, leading to reduced efficiency and accuracy. In contrast, applications and systems can improve the accuracy of their decisions and actions based on high-quality sensing data, helping them perform more efficiently and effectively. The impact of low-quality data highlights the importance of evaluating and enhancing the quality of the sensing data. Since the sensing data are measured and submitted by MUs, the sensing platform lacks the ground truth of these data, making it difficult to directly evaluate their quality. Consequently, considerable effort has been devoted to estimating the ground truth of sensing data, leading to the application of three types of methods in MCS systems: data aggregation methods \cite{rst,decentralized,MAB,onlineQ,estimateQ}, prediction-based methods \cite{starima,sttn,MVSTGN,glsttn,stresnet,series,kalman,mcspredict,lowmalicious}, and truth discovery (TD) methods \cite{encrytd,lighttd,sctd,mcstd,pptd,tf,epptd,dtd,tffds,generaltf}. Data aggregation methods estimate the ground truth of sensing data through clustering algorithms. However, their estimates may be inaccurate when insufficient sensing data are available \cite{onlineQ,estimateQ}. To address this problem, several schemes have designed MCS frameworks with pre-exploration stages to obtain sufficient baseline sensing data \cite{MAB,onlineQ}. Unfortunately, these methods increase the overhead associated with MCS. As many sensing tasks in MCS applications are repeated periodically or continue for a very long period, historical data can be obtained in many cases and are useful for ground truth estimation. Prediction-based methods initially use these historical data to learn and predict the ground truth of the sensing data submitted by MUs, subsequently calculating the errors between these data and the corresponding predicted ground truth to evaluate the data quality. On this basis, Xie \textit{et al}. \cite{sttn} and Zhang \textit{et al}. \cite{stresnet} attempted to capture the spatial and temporal correlations among data and used them to predict the ground truth. However, these methods may not perform well in realistic MCS systems with sensing data, including bursty values.

Numerous schemes have applied TD in quality-based MCS systems to evaluate the quality of conflicting data \cite{encrytd,lighttd,sctd,mcstd}. TD methods initially aggregate the sensing data submitted by MUs based on their weights (i.e., trustworthiness levels) to estimate the ground truth (e.g., calculate the weighted mean of all submitted data as the ground truth). The weights of the MUs are subsequently updated based on the errors between their submitted data and the estimated ground truth. Specifically, the weights of MUs are increased if they provide sensing data that are close to the estimated ground truth, resulting in greater impacts on the estimation process. The ground-truth estimates subsequently need to be calculated again as the weights of the MUs are adjusted. Therefore, the weights of the MUs and the ground-truth estimates are alternately updated until convergence is reached. To further improve the framework of TD methods, Cheng \textit{et al}. \cite{pptd} adopted a reputation mechanism \cite{design_rep,reputation,bcreput} for TD in an MCS system to replace the weights of MUs with reputations. This approach effectively reduces the misclassification of MUs. Moreover, Yin \textit{et al}. \cite{tf} utilized the implications among conflicting data and designed a framework based on TD to identify incorrect and fake data. However, the effectiveness of TD methods is limited by the quantity of the given sensing data. Like the data aggregation methods, TD methods estimate ground truth with low accuracy in scenarios where the sensing data are insufficient and sparse.

Note that in MCS systems, the sensing data submitted by MUs exhibit spatial and temporal correlations \cite{pp-assign, sctd, corre, stresnet} in most scenarios (e.g., the traffic flows of a city and the temperature and humidity of different regions), which can be utilized to improve the performance of the data quality evaluation. On this basis, we propose a prediction- and reputation-based TD (PRBTD) framework to evaluate the quality of sensing data with spatio-temporal correlations.

The proposed PRBTD method leverages the advantages of prediction-based methods, TD methods, and reputation mechanisms. First, we adopt a correlation-focused spatio-temporal Transformer network (CFSTTN), which is an improved version of the global-local spatio-temporal Transformer network (GLSTTN) \cite{glsttn}, to learn from historical sensing data and predict the ground truth of the submitted sensing data. The CFSTTN incorporates a long-range residual connection to enhance its ability to extract the spatio-temporal correlations among data. However, in practice, the prediction results may be inaccurate because of the noise in historical sensing data and the bursty data values. Therefore, instead of directly using the prediction results for data quality evaluation, we extract the implications among data at different locations and in different periods based on the prediction results, which can remain stable and thus reduce the impact of inaccurate predictions. Finally, our PRBTD framework identifies malicious MUs and low-quality data through a TD strategy based on the reputations of the MUs and the implications among the data. Thus, our proposed method enhances the quality of the sensing data by filtering out incorrect and fake data. The main contributions of this work can be summarized as follows.

\begin{itemize}
	\item The proposed PRBTD framework consists of a CFSTTN prediction module, a data feature and implication calculation module, and a reputation-based TD module. This approach combines prediction, TD, and reputation mechanisms for the first time to evaluate the quality of the sensing data and is applicable in most scenarios, including cases with bursty data values and data sparsity.
	\item The data feature and implication calculation module in PRBTD uses the prediction results to analyze the implications among the data acquired from various locations and in different periods. Since these implications remain stable, PRBTD reduces the impact of prediction inaccuracies by focusing on implications instead of directly using the prediction results. Therefore, PRBTD overcomes the shortcomings of prediction-based methods, especially in scenarios with bursty sensing data values.
	\item The reputation-based TD module in PRBTD leverages the implications among the data and the reputations of MUs to achieve accurate evaluation of data quality. By utilizing these implications, the reputation-based TD module indirectly increases the amount of data available for evaluating the quality of the submitted data. Consequently, PRBTD can more accurately evaluate data quality in scenarios with sparse sensing data.
	\item We conduct extensive simulation experiments on real-world datasets. The results indicate that our PRBTD framework is applicable to complex scenarios in MCS systems and outperforms existing methods.
\end{itemize}

\section{Related Work}
In this section, we briefly review the related work on quality-based MCS systems and methods for evaluating data quality.

\subsection{Quality-Based MCS Systems}
Numerous studies have been dedicated to the establishment of data quality-based MCS systems \cite{qoi,macnb,cmabb,olqa,optim}. An effective approach is to design incentive and task assignment mechanisms that induce MUs to submit high-quality data. For example, Jin \textit{et al.} \cite{qoi} designed an incentive mechanism based on reverse combinatorial auctions, aiming to encourage the participation of MUs and maximize social welfare. Xiao \textit{et al.} \cite{cmabb} used a multi-armed bandit algorithm to select MUs with good performance to ensure data quality. Gu \textit{et al.} \cite{macnb} adopted a Stackelberg leader-follower game framework in MCS systems to design a task assignment mechanism that is able to choose MUs with high potential through multiagent reinforcement learning. However, these methods assume prior data quality knowledge, which does not always hold in practice. Therefore, some methods have added pre-exploration stages to MCS systems to estimate the capabilities of MUs \cite{olqa}. For example, Zhang \textit{et al.} \cite{onlineQ} and Gao \textit{et al.} \cite{MAB} used the data sensed in the pre-exploration stages as baselines for evaluating the sensing errors of MUs, which could represent their abilities. Although these methods can more accurately evaluate the quality of the sensing data, the overhead of the resulting MCS platform is greatly increased.

\subsection{Data Quality Evaluation}
Many studies have focused on evaluating the quality of the data collected within MCS systems \cite{taskme,corre}. Zhang \textit{et al.} \cite{onlineQ} and Jin \textit{et al.} \cite{qoi} compared sensing data with the corresponding ground truth to calculate the induced sensing errors and to evaluate the quality of the data. On the basis of these results, Gao \textit{et al.} \cite{MAB} decomposed the sensing errors into deviations and variances to represent the sensing abilities and stabilities of MUs. However, these methods assume that the ground truth is known to the sensing platform, which is impractical in MCS systems. Therefore, many studies have been conducted on ground truth estimation, resulting in three types of methods: data aggregation methods \cite{rst,decentralized,estimateQ,MAB,onlineQ}, prediction-based methods \cite{starima,sttn,MVSTGN,glsttn,stresnet,series,kalman,mcspredict,lowmalicious}, and TD methods \cite{encrytd,lighttd,sctd,mcstd,pptd,tf,epptd,dtd,tffds}.

\textbf{Data aggregation methods.} 
Zhang \textit{et al.} \cite{onlineQ} applied a weighted averaging method to estimate the ground truth of the sensing data. However, this method requires sufficient sensing data to obtain accurate estimates, which is difficult to achieve in scenarios with sparse data. To address data sparsity, Meng \textit{et al.} \cite{rst} introduced a method for estimating the quality of the sensing data via a sparse data matrix. Specifically, they constructed a data matrix from submitted data and then filled in the missing data via Bayesian probabilistic matrix factorization. Building on this contribution, Yang \textit{et al.} \cite{decentralized} deployed a small number of unmanned aerial vehicles to collect trustworthy data for constructing the sparse baseline data matrix. However, these methods focus only on the spatial correlations among data and overlook the temporal aspect, which limits their effectiveness.

\textbf{Prediction-based methods.} Min \textit{et al.} \cite{starima} utilized the spatio-temporal correlations among traffic data and achieved improved prediction performance with their proposed dynamic space-time autoregressive integrated moving average (STARIMA) model. Furthermore, Yao \textit{et al.} \cite{MVSTGN} applied a Transformer framework for traffic prediction and proposed a novel multi-view spatio-temporal graph network. This model combines attention and convolution mechanisms in its traffic pattern analysis strategy to comprehensively determine the spatio-temporal characteristics. On the basis of this approach, Gu \textit{et al.} \cite{glsttn} proposed a Transformer model with global and local spatio-temporal modules to analyze the strong dynamics and spatio-temporal dependencies that are inherent in cellular traffic data. However, the predictions yielded by these methods may be inaccurate when bursty values are present in the sensing data, which limits the performance of prediction-based methods in MCS systems.

\textbf{TD methods.} TD methods are considered efficient for estimating the ground truth from unreliable data \cite{encrytd,lighttd,sctd,mcstd} and are thus suitable for MCS systems. For example, Cheng \textit{et al.} \cite{pptd} presented a framework that combines TD with zero-knowledge proof in an MCS system to improve the accuracy of the ground truth produced for sensing data while preserving user privacy. Yin \textit{et al.} \cite{tf} leveraged TD to evaluate the trustworthiness of web-based information sources and identify incorrect and fake data. However, these traditional TD methods typically produce unsatisfactory results in practical applications because of the possibility of sparse sensing data being present in MCS systems.

In summary, most existing methods can handle only specific scenarios. Inspired by these works, we have found that prediction-based methods and TD methods can effectively complement each other. Therefore, we propose leveraging the principles of prediction-based methods and reputation-based TD methods while fully utilizing the spatio-temporal correlations among data to implement our PRBTD method, which is able to enhance the effectiveness of data quality evaluation across various complex scenarios.

\section{MCS System Model}
\begin{figure}[!t]
	\centering{\includegraphics[width=\columnwidth]{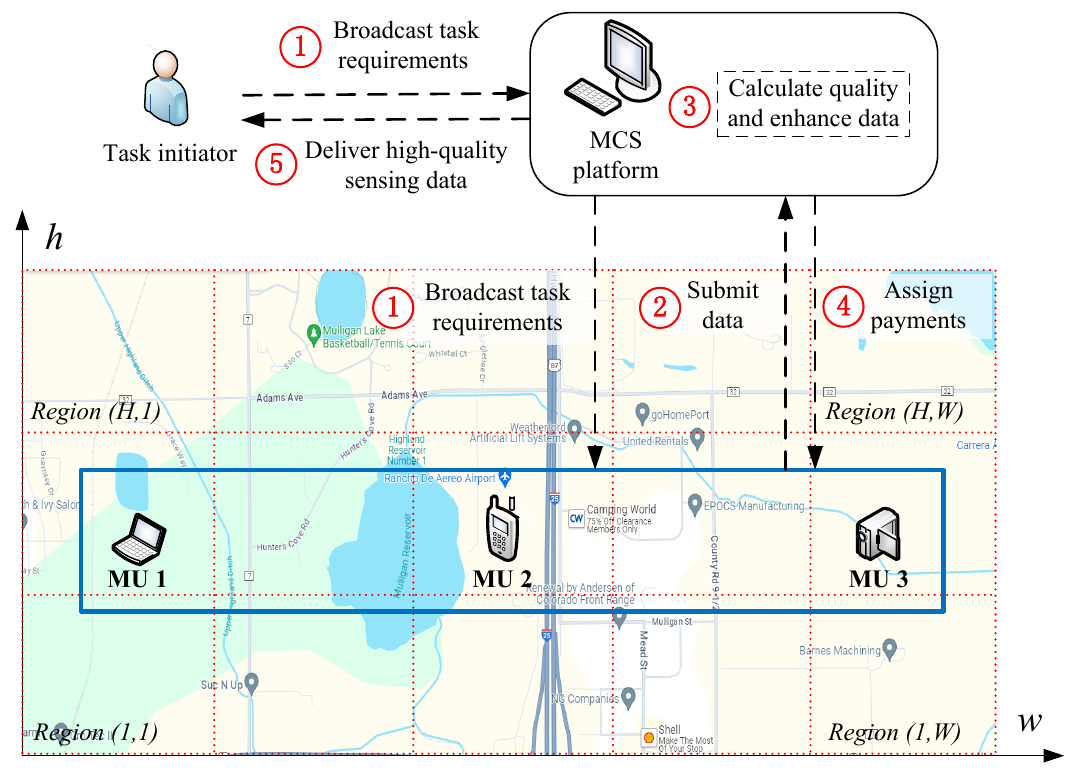}}
	\caption{Architecture of an MCS system.}
	\label{fig-mcs}
\end{figure}
The architecture of an MCS system is illustrated in \mbox{Fig. \ref{fig-mcs}}. In MCS, a sensing task typically spans a duration of $T$ time slots and covers an area with $N$ regions \cite{corre, dtd}. In this work, we study scenarios with historical sensing data, and these data collected in a task exhibit temporal and spatial correlations (e.g., collecting traffic flow data in a city for several days). The time slot of the task is denoted by $t\in\left\{1,2,...,T \right\}$. The $N(=H\times W)$ sensing regions are identified based on their latitudes and longitudes. To simplify the expression, we use $n \in \left\{1,2,...,N\right\}$ to identify a region ($h,w$), where $n=(h-1) \times W + w$. An MU in the MCS system is identified as $i\in\left\{1,2,...,I \right\}$, where $I$ is the total number of MUs. In particular, we assume that each MU submits sensing data at most once per time slot. The data submitted to the platform by an MU should include the ID $i$ of the MU, the sensing time slot $t$, the sensing location $n$, and the sensing value $v_{i,n}^t$. In our systems, we assume that the sensing data have continuous data values. However, our proposed PRBTD method can be generalized to other data types (e.g., categorical sensing data). Thus, we represent the sensing data point submitted by MU $i$ in region $n$ at time slot $t$ as $d_{i,n}^t=\left(i,t,n,v_{i,n}^t\right)$, and use $\bm{D}_n^t$ to denote the set of all the data sensed in region $n$ at time slot $t$. Moreover, there are malicious MUs in our system. Malicious MUs are users who continuously submit low-quality data that are harmful to the system. Similar to \cite{tf, lowmalicious}, we assume that the proportion of malicious MUs in our system should not exceed 50\%. This assumption is based on the expectation that the majority of MUs are honest \cite{tf}. If the proportion of malicious MUs exceeds 50\%, the system can be considered under attack, making the system untrustworthy \cite{reputation}. In such a case, the majority of the sensing data may be malicious, rendering most existing methods for data quality evaluation ineffective. The interactions among the TI, MUs, and sensing platform in the MCS system are as follows \cite{dtd, macnb}.

\begin{enumerate}
	\item At the beginning of a sensing task, the TI broadcasts the task requirements to the available MUs through the sensing platform.
	\item During each time slot of the task, the MUs may choose to sense and submit their data to the sensing platform. At the end of each time slot, the sensing platform gathers and identifies the incorrect or fake data among all submitted data. These low-quality data are then eliminated by the platform to enhance the quality of the remaining data.
	\item At the end of the task, payments are assigned to the MUs based on the quality of their sensing data, and high-quality data are delivered to the TI.
\end{enumerate}

\section{Our Method}
\begin{figure}[!t]
	\centering{\includegraphics[width=\columnwidth]{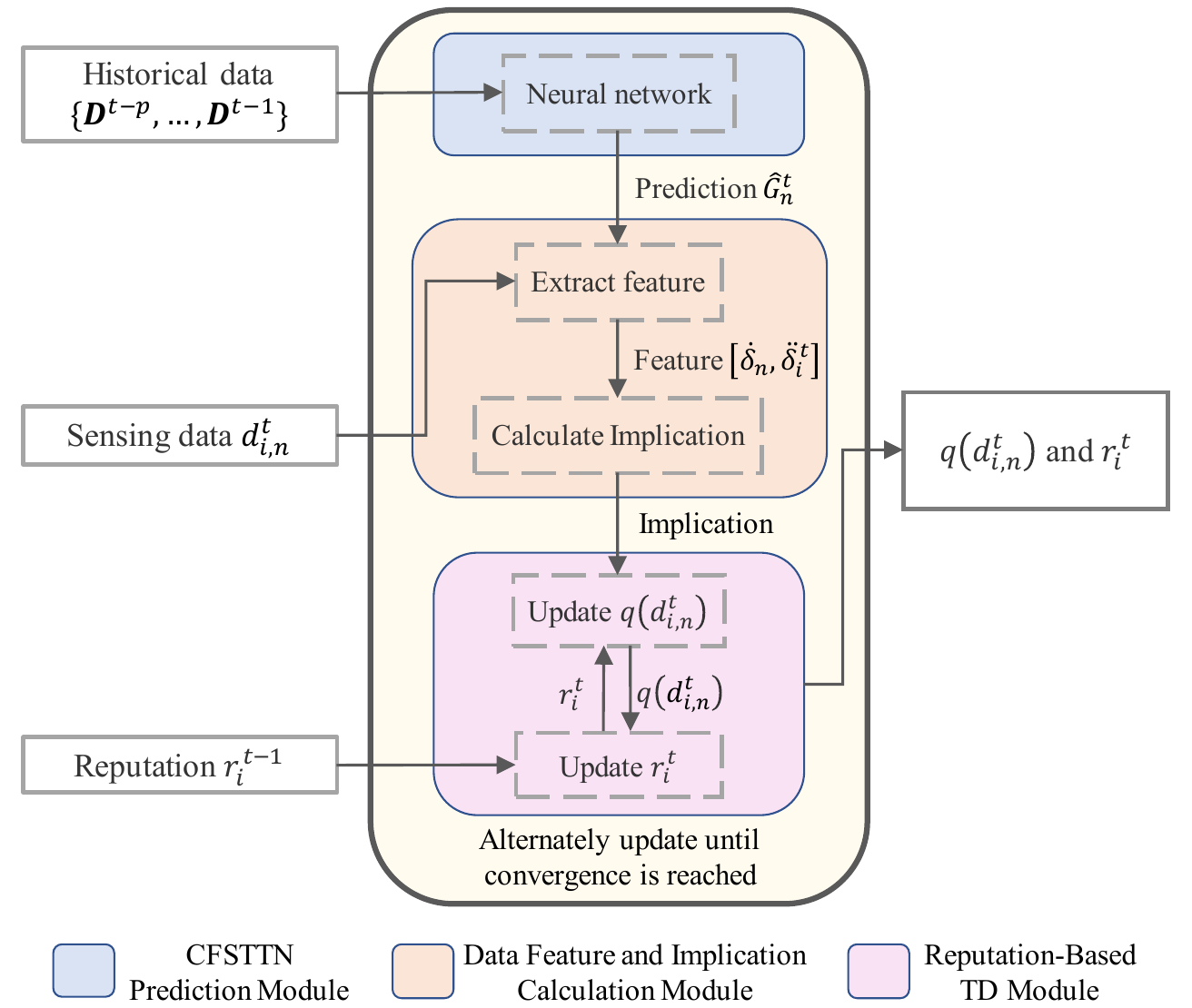}}
	\caption{Architecture of the PRBTD framework.}
	\label{fig-prbtd}
\end{figure}

In this section, we introduce the proposed PRBTD framework. As shown in Fig. \ref{fig-prbtd}, our PRBTD framework consists of three modules: a CFSTTN prediction module, a data feature and implication calculation module, and a reputation-based TD module. Table \ref{table-notation} lists the notations used in the PRBTD framework.

\begin{table}[!t]
	\caption{Notations Used in the PRBTD Framework}
	\label{table-notation}
	\centering
	\setlength{\tabcolsep}{3pt}
	\renewcommand{\arraystretch}{1.3}
	\begin{tabular}{p{60pt}|p{180pt}}
		\hline
		Notation&
		Description\\
		\hline
		$I, i$&
		The total number of MUs, the ID of an MU\\
		$T, t$&
		The total number of time slots for a sensing task, a time slot in the task duration\\
		$N, n$&
		The total number of regions in a sensing task, a region in the task area\\
		$d_{i,n}^t$&
		The sensing data submitted by MU $i$ for region $n$ at time slot $t$\\
		$v_{i,n}^t$&
		The value of sensing data point $d_{i,n}^t$\\
		$\bm{I}(v_{i,n}^t)$&
		The set of MUs who sense the same value $v_{i,n}^t$ in region $n$ at time slot $t$\\
		$\bm{D}_n^t$&
		The set of data sensed in region $n$ at time slot $t$\\
		$\bm{D}^{\prime}, \widehat{\bm{D}}^t$&
		The data buffer, the set of high-quality data at time slot $t$\\
		$M_n^t, \bm{M}_n$&
		The mean of the sensing data values in region $n$ at time slot $t$, the set of mean data values at time slot $t$ in all regions\\
		$\widehat{G}_n^t, \widehat{\bm{G}}^t$&
		The ground-truth predictions of data in region $n$ at time slot $t$, the set of ground-truth predictions at time slot $t$\\
		$\delta_{i,n}^t$&
		The sensing error of data point $d_{i,n}^t$\\
		$\dot{\delta}_n^t, \ddot{\delta}_i^t$&
		The circumstance-related error of region $n$ at time slot $t$, the user-related error of MU $i$ at time slot $t$\\
		$r_i^t, R_i^t$&
		The reputation of MU $i$ at time slot $t$, the corresponding reputation score\\
		$q(d_{i,n}^t), Q(d_{i,n}^t)$&
		The data quality of sensing data point $d_{i,n}^t$, the corresponding quality score\\
		$imp\!\left(d_{i,n}^t, d_{j,n^{\prime}}^{t^{\prime}}\right)$&
		The implication between sensing data points $d_{i,n}^t$ and $d_{j,n^{\prime}}^{t^{\prime}}$\\
		\hline
	\end{tabular}
\end{table}

\subsection{CFSTTN Prediction Module}
In an MCS system, the sensing platform cannot directly evaluate the quality of the submitted data due to the absence of the ground truth. However, in scenarios with historical sensing data, prediction methods can be employed to estimate the ground truth. Thus, we propose a CFSTTN to learn from historical data and predict the ground truth of the submitted sensing data.

\begin{figure*}[!t]
	\centering{\includegraphics[width=\linewidth]{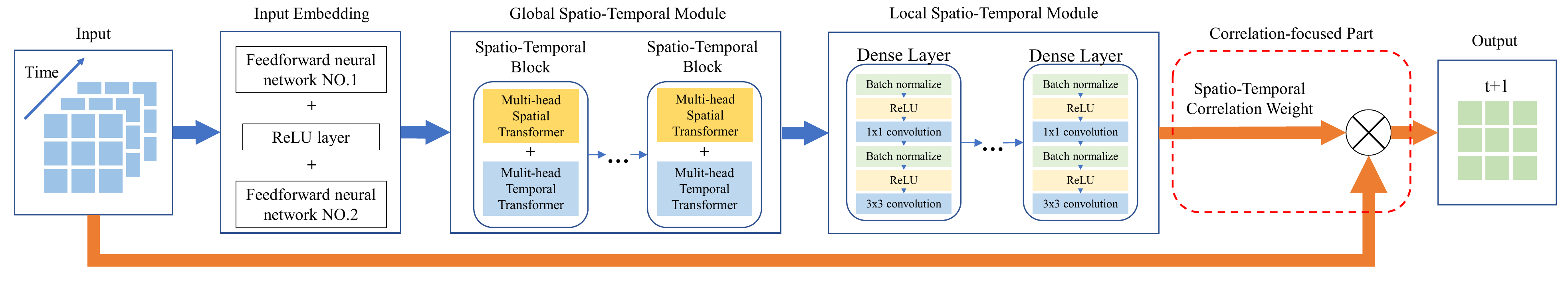}}
	\caption{Structure of the CFSTTN.}
	\label{fig-cfsttn}
\end{figure*}

As shown in Fig. \ref{fig-cfsttn}, the CFSTTN comprises three components: an embedding module, a global spatio-temporal module, and a local spatio-temporal module. First, the embedding module adjusts the dimensions of the input data. Then, the global spatio-temporal module deploys a spatio-temporal Transformer framework based on a multi-head attention mechanism to extract multidimensional global spatial and temporal characteristics from the input data. Finally, the local spatio-temporal module uses densely connected convolutional neural networks to further extract local spatio-temporal information based on the global characteristics of the input data.

Our proposed CFSTTN improves the original model by integrating a correlation-focused part with a long-range residual connection with the input data. As presented in Fig. \ref{fig-cfsttn}, our CFSTTN predicts a spatio-temporal relationship weight matrix, which is subsequently multiplied with the input data to derive the final output. The integration of the long-range residual connection enhances the spatio-temporal correlations between the output and input data, thus improving the prediction accuracy.

The CFSTTN uses historical data in the sensing task for prediction purposes. We represent the mean values of the data at time slot $t$ as a column vector $\bm{M}^t = \left[M_1^t, M_2^t,...,M_N^t\right]^{\mathsf{T}}$ where
\begin{equation}\label{eqM}
	M_n^t = \frac{\sum_{d_{i,n}^t \in \bm{D}_n^t} v_{i,n}^t}{|\bm{D}_n^t|}
\end{equation}
is the mean value of the data sensed at time slot $t$ in region $n$. We construct a matrix $\left[\bm{M}^{t-p+1} ,...,\bm{M}^t\right] \in \mathbb{R}^{N \times p}$ as the input of the CFSTTN, which consists of the mean values of data derived from the preceding $p$ time slots up to $t$, where $0<p<T$ is a hyperparameter. The predicted ground truth, which includes the mean values of the data at time slot $t+1$ from every region, is denoted as $\widehat{\bm{G}}^{t+1} = \left[\widehat{G}_1^{t+1},\widehat{G}_2^{t+1},...,\widehat{G}_N^{t+1}\right]^{\mathsf{T}}$. Additionally, we use $f(\cdot)$ to represent a series of operations conducted within the three components of the CFSTTN. The spatio-temporal correlation weight matrix can be represented as $f(\bm{M}^{t-p+1} ,...,\bm{M}^t)$. Finally, we obtain the prediction of the ground truth at time slot $t+1$ by performing matrix multiplication between the spatio-temporal correlation weight matrix and the input data:
\begin{equation}\label{eq-predict}
	\widehat{\bm{G}}^{t+1} = \left[\bm{M}^{t-p+1} ,...,\bm{M}^t\right]f(\bm{M}^{t-p+1} ,...,\bm{M}^t).
\end{equation}

In this paper, the mean absolute error (MAE) is utilized as the training metric. Specifically, we use the backpropagation method to minimize the MAE between the predicted ground truth $\widehat{\bm{G}}^{t+1}$ and the mean values of the actual submitted data $\bm{M}^{t+1}$, which can be expressed as follows:
\begin{equation}\label{eq-loss}
	\mathcal{L}\left(\theta\right) = \frac{1}{N}\sum_{n=1}^{N}\left|\widehat{G}_n^{t+1}-M_n^{t+1}\right|
\end{equation}
where $\theta$ represents the parameters in the CFSTTN, which are learned and updated during the training process. However, the prediction results in an MCS task may be inaccurate because of the noise in the training data and the bursty values in the sensing data. Therefore, the output of the CFSTTN prediction module cannot be directly used to evaluate the quality of the data.

\subsection{Data Feature and Implication Calculation Module}
\begin{figure}[!t]
	\centering{\includegraphics[width=0.9\columnwidth]{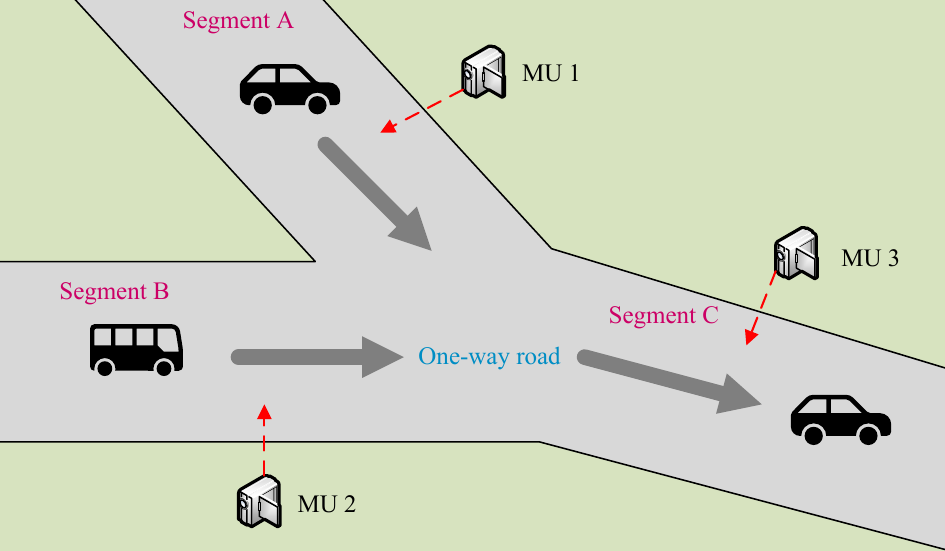}}
	\caption{Simple example of MUs sensing traffic flow data.}
	\label{fig-imp}
\end{figure}

In the MCS system, implicit correlations are present among the sensing data with spatial and temporal correlations. We denote the implicit correlation among the sensing data points as $implication$, which reflects the extent to which one data point can predict or support another. For example, consider the scenario shown in Fig. \ref{fig-imp} where MUs collect traffic flow data on a one-way road. In this case, the traffic flow on segment C is the sum of the traffic flows on segment A and segment B at any period $t$. Therefore, when the traffic data points submitted by MUs satisfy $v_{1,A}^t + v_{2,B}^t = v_{3,C}^t$, the implications among these data points are strong, as they mutually support each other's accuracy. In contrast, the implications are weak when the data points follow $v_{1,A}^t + v_{2,B}^t \neq v_{3,C}^t$, which means that some of these MUs have submitted incorrect data that thus contradict the other data. Similarly, in most MCS scenarios, we can infer the existence of implications among the sensing data. However, owing to the complex spatial and temporal correlations among the data, quantitatively analyzing these implications is challenging in most cases. Therefore, we attempt to understand these implications from historical data. Specifically, we perform a series of computations in this module to quantify these implications from the output of the CFSTTN prediction module. Before providing the detailed calculations, we define the degree of implication as follows:

\begin{definition} \label{def1}
	(Degree of implication) The degree of implication between data points $d_{i,n}^t$ and $d_{j,n^\prime}^{t^\prime}$ is represented as $imp(d_{i,n}^t,d_{j,n^\prime}^{t^\prime})$, where $n^\prime \in \left\{1,2,...,N\right\}$ and $0<t^\prime \leq t \leq T$. Moreover, the degree satisfies $-1 \leq imp(d_{i,n}^t,d_{j,n^\prime}^{t^\prime}) \leq 1$.
\end{definition}

Specifically, $imp(d_{i,n}^t,d_{j,n^\prime}^{t^\prime})\!=\!0$ indicates a lack of implication between data points $d_{i,n}^t$ and $d_{j,n^\prime}^{t^\prime}$, whereas $imp(d_{i,n}^t,d_{j,n^\prime}^{t^\prime})>0$ indicates mutual support between the data, and $imp(d_{i,n}^t,d_{j,n^\prime}^{t^\prime})<0$ means that the data contradict each other. Implications close to 1 indicate that the data points strongly support each other, whereas implications close to -1 suggest that the data points are likely to be entirely contradictory.

\begin{figure*}[!t]
	\centering
	\begin{minipage}[b]{0.85\columnwidth}
		\includegraphics[width=\linewidth]{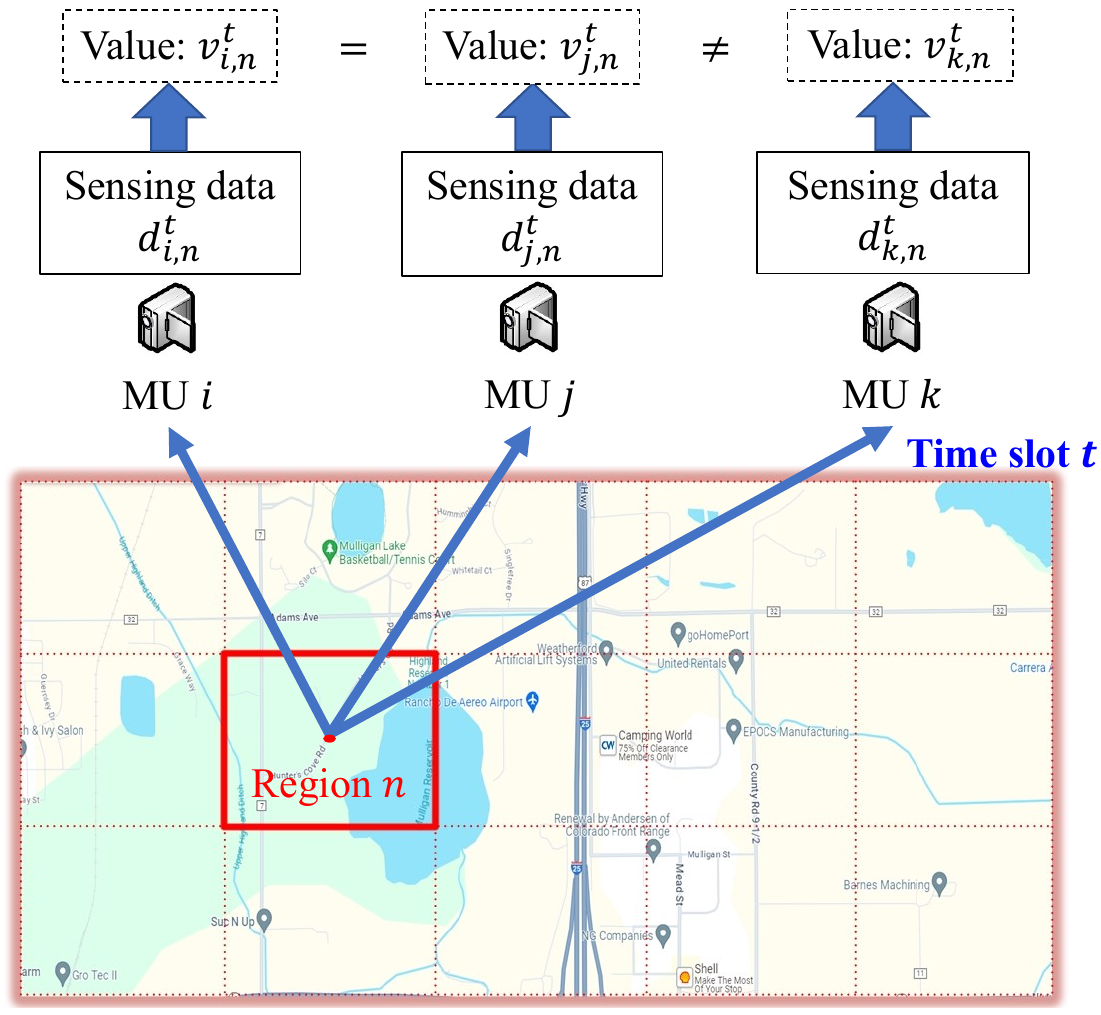}
		\centering {\fontsize{8pt}{9pt}\selectfont (a) MUs sense data in the same region at the same time slot.}
		\label{fig-scene-b}
	\end{minipage}
	\hspace{12mm}
	\begin{minipage}[b]{0.9\columnwidth}
		\includegraphics[width=\linewidth]{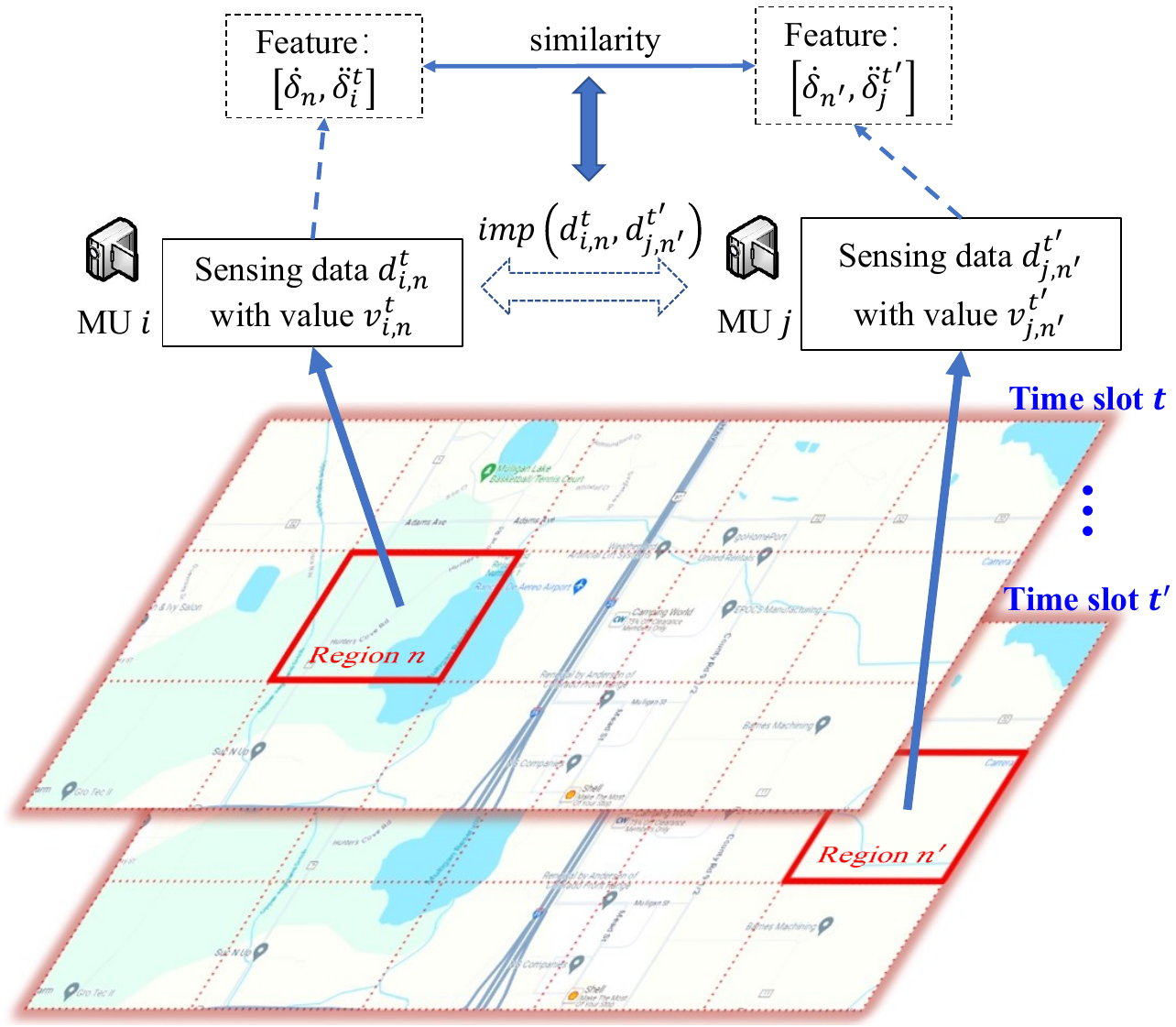}
		\centering {\fontsize{8pt}{9pt}\selectfont (b) MUs sense data in different regions at different time slots.}
		\label{fig-scene-a}
	\end{minipage}
	\hspace{-8mm}
	\caption{MCS scenarios with multiple MUs.}
	\label{fig-scene}
\end{figure*}

To calculate the degree of implication between sensing data points, we first reduce the dimensionality \cite{pca} of the sensing data $d_{i,n}^t=\left(i,t,n,v_{i,n}^t\right)$ to extract the features. Specifically, since sensing errors can reflect the characteristics of data \cite{onlineQ,MAB}, we calculate the sensing errors as data features. Since the actual ground truth of the sensing data are unknown to us during a sensing task, we calculate the sensing errors of the data points based on the ground truth prediction results (i.e., the output of the CFSTTN). When the corresponding predicted ground truth $\widehat{G}_n^t \neq 0$, the error of data point $d_{i,n}^t$ is calculated as
\begin{equation}\label{eq-delta}
	\delta_{i,n}^t = \frac{v_{i,n}^t - \widehat{G}_n^t}{\widehat{G}_n^t}.
\end{equation}

In most MCS scenarios, the sensing errors in the data are caused primarily by two factors \cite{MAB}. The first factor is related to external sensing circumstances, such as weather conditions, which are influenced by the time and location of the data collection. The second factor is related to the MUs, such as issues with the devices or improper operations when sensing data. Therefore, we decompose the sensing errors in the data into two components, which are defined below.
\begin{definition}\label{c-error}
	(Circumstance-related error) The error caused by external factors (e.g., the illumination intensity or weather conditions) during data collection is denoted as $\dot{\delta}_n^t$. This error is influenced by the sensing region and period.
\end{definition}
\begin{definition}\label{u-error}
	(User-related error) The error caused by internal factors (e.g., devices or operations) of MU $i$ during data collection is denoted as $\ddot{\delta}_i^t$. Note that $\ddot{\delta}_i^t$ is also related to the sensing period $t$, as the internal factors of an MU may vary in different periods.
\end{definition}
Hence, the total sensing error of a data point $d_{i,n}^t$ can be divided into the circumstance-related error and the user-related error as follows:
\begin{equation}\label{eq-decompose}
	\delta_{i,n}^t = \dot{\delta}_n^t + \ddot{\delta}_i^t
\end{equation}

On this basis, we learn that the error vector $[\dot{\delta}_n^t, \ddot{\delta}_i^t]$ corresponds to, and is uniquely related to, the data point $d_{i,n}^t$. Therefore, we use the error vector of a data point as its data feature, which can reflect most of the characteristics of that data point. Specifically, we can analyze the relationship between the features of the data points and the implications among them through the following conjecture, which allows us to infer the implications among data points based on their features.

\begin{conjecture}\label{conj1}
	The degree of implication between two data points is positively correlated with the similarity between the corresponding data features \cite{tf,generaltf}.
\end{conjecture}

This conjecture is applicable in most MCS scenarios. The reason is that data features reflect their patterns; thus, data with similar features usually follow a similar pattern and can infer each other. In our work, the features we extract from the sensing data include both circumstance-related errors and user-related errors. Therefore, if two data points have similar features, their circumstance-related errors and user-related errors are similar. According to Definitions \ref{c-error} and \ref{u-error}, these two data points may be collected by users with similar characteristics (e.g., using similar devices or similar operations) in similar circumstances (e.g., under similar weather conditions). In such cases, the two data points can strongly support each other's authenticity \cite{tffds}. For example, when traffic flow data are sensed on different road segments with similar traffic characteristics (e.g., parallel roads with comparable traffic flows), if the data from two segments are very close, they can mutually support each other. Hence, the implication between two data points with similar features is strong. In contrast, data points with dissimilar features may contradict each other; therefore, the implications among these data are weak. Consequently, we can infer that in most MCS scenarios, Conjecture \ref{conj1} is reasonable.

Based on Conjecture \ref{conj1}, we calculate the cosine similarity of the data features to determine the degree of implication between the data points as shown in Fig. \ref{fig-scene}(b). Even though the cosine similarity focuses only on direction, it is an appropriate choice for calculating the similarity of data features because our data features already contain magnitude information. The detailed process of calculating the data features and degrees of implication is as follows.

\textbf{Calculate the circumstance-related error $\bm{\dot{\delta}_n^t}$.} In our MCS system, the proportion of malicious MUs is low \cite{tf, reputation, lowmalicious}, and the MCS platform can identify and exclude these malicious MUs during sensing tasks. Therefore, we assume that the impact of malicious MUs on the calculations is limited. 

According to \eqref{eq-decompose}, we calculate the average sensing error of all the data collected in region $n$ as follows:
\begin{equation}\label{eq-average}
	\frac{\sum_{\tau=1}^{\infty} \sum_{d_{i,n}^{\tau}\in\bm{D}_{n}^{\tau}} \delta_{i,n}^{\tau}}{\sum_{\tau=1}^{\infty} |\bm{D}_{n}^{\tau}|} = \frac{\sum_{\tau=1}^{\infty} \sum_{d_{i,n}^{\tau}\in\bm{D}_{n}^{\tau}} \left(\dot{\delta}_n^{\tau}+\ddot{\delta}_i^{\tau}\right)}{\sum_{\tau=1}^{\infty} |\bm{D}_{n}^{\tau}|}
\end{equation}
where $\bm{D}_{n}^{\tau}$ represents the set of data sensed in region $n$ at time slot $\tau$. We use $c_n^t$ to represent the amount of sensing data collected in region $n$ up until time slot $t$. Specifically, $c_n^t$ is calculated as follows:
\begin{equation}\label{eq-cnt}
	c_n^t = \begin{cases}
		0, & t=0\\
		c_n^{t-1} + |\bm{D}_{n}^t|, & 0<t\leq T.
	\end{cases}
\end{equation}
Based on \eqref{eq-cnt}, we can rewrite \eqref{eq-average} as follows:
\begin{equation}\label{eq-relation}
	\sum_{\tau=1}^{\infty} \left(c_n^\tau - c_n^{\tau-1}\right) \dot{\delta}_n^\tau = 
	\sum_{\tau=1}^{\infty} \sum_{d_{i,n}^{\tau}\in\bm{D}_{n}^{\tau}} \delta_{i,n}^{\tau}.
\end{equation}
Note that when simplifying \eqref{eq-average} to \eqref{eq-relation}, we utilize the property that the user-related error of each normal MU $i$ satisfies
\begin{equation}\label{eq-sumdelta}
	\sum_{\tau=1}^{\infty} \ddot{\delta}_i^{\tau} = 0.
\end{equation}
The reason is that the long-term user-related error expectation for each normal MU in an MCS system should be zero.

However, the relationship between $\dot{\delta}_n^t$ and $\delta_{i,n}^t$ shown in \eqref{eq-relation} holds only as $t\to\infty$, which is not feasible in practice because the duration of a sensing task is limited. Therefore, we rewrite \eqref{eq-relation} in the following slot-based update form to calculate the estimated circumstance-related error $\dot{\delta}_n^{(t)}$ in region $n$ at time slot $t$:
\begin{equation}\label{eq-cre}
	\dot{\delta}_n^{(t)} = \begin{cases}
		0, &c_n^t=0\\
		\frac{\dot{\delta}_n^{(t-1)}c_n^{t-1}+\sum_{d_{i,n}^t\in\bm{D}_n^t} \delta_{i,n}^t}{c_n^t}, &c_n^t\neq0
	\end{cases}
\end{equation}
Note that $\lim_{t \to \infty}\dot{\delta}_n^{(t)} = \dot{\delta}_n^t$. Thus, a longer sensing task duration and more recent time slots can result in more accurate estimates. Therefore, we use $\dot{\delta}_n^{(t)}$ calculated from \eqref{eq-cre} instead of $\dot{\delta}_n^t$ in the following calculations.

\textbf{Calculate the user-related error $\bm{\ddot{\delta}_i^t}$.} Suppose we have a case where $\widehat{G}_n^t \neq 0$, then $\ddot{\delta}_i^t$ can be calculated according to \eqref{eq-delta} and \eqref{eq-decompose} as follows:
\begin{equation}\label{eq-ure}
	\ddot{\delta}_i^t = \frac{v_{i,n}^t - \left(1+\dot{\delta}_n^{(t)}\right)\widehat{G}_n^t}{\widehat{G}_n^t}.
\end{equation}

However, the calculation in \eqref{eq-ure} neglects the situation in which $\widehat{G}_n^t = 0$. Considering that the cosine similarity between vectors remains unchanged when the vectors are scaled, we scale the error vector $[\dot{\delta}_n^{(t)},\ddot{\delta}_i^t]$ of data point $d_{i,n}^t$ by $\widehat{G}_n^t$ to form a new data feature $F_{i,n}^t = [\widehat{G}_n^t\dot{\delta}_n^{(t)}, v_{i,n}^t - (1+\dot{\delta}_n^{(t)})\widehat{G}_n^t]$. The scaled feature $F_{i,n}^t$ can be utilized to calculate the degree of implication, and the calculation process remains valid even when $\widehat{G}_n^t=0$.

\textbf{Calculate the degree of implication between sensing data points $\bm{d_{i,n}^t}$ and $\bm{d_{j,n^{\prime}}^{t^\prime}}$.} We calculate the cosine similarity between the data feature vectors to determine the degree of implication as follows:
\begin{equation}\label{eq13}
	\begin{aligned}
		imp(d_{i,n}^t,d_{j,n^\prime}^{t^\prime}) &= cosine\ similarity(F_{i,n}^t, F_{j,n^\prime}^{t^\prime})\\
		&=\frac{F_{i,n}^t \cdot F_{j,n^\prime}^{t^\prime}}{\|F_{i,n}^t\| \|F_{j,n^\prime}^{t^\prime}\|}
	\end{aligned}
\end{equation}
where $F_{i,n}^t$ is the scaled feature vector of the sensing data point $d_{i,n}^t$ and $F_{j,n^\prime}^{t^\prime} = [\widehat{G}_{n^\prime}^{t^\prime}\dot{\delta}_{n^\prime}^{(t)}, v_{j,n^\prime}^{t^\prime} - (1+\dot{\delta}_{n^\prime}^{(t)})\widehat{G}_{n^\prime}^{t^\prime}]$ corresponds to the data point $d_{j,n^{\prime}}^{t^\prime}$. Note that when we calculate $imp(d_{i,n}^t,d_{j,n^\prime}^{t^\prime})$ according to \eqref{eq13}, the circumstance-related error of region $n^{\prime}$ is $\dot{\delta}_{n^\prime}^{(t)}$ instead of $\dot{\delta}_{n^\prime}^{(t^\prime)}$. Since the time slots satisfy $t^\prime \leq t$ according to Definition \ref{def1}, indicating that time slot $t$ is more recent than the other time slots are, the estimated circumstance-related error of region $n^\prime$ should be updated to $\dot{\delta}_{n^\prime}^{(t)}$, which is more accurate.

Finally, the implication between any two data points can be extracted and utilized for the calculations in the subsequent module. Even in scenarios with bursty sensing data values, the implicit spatial and temporal correlations between data points remain stable \cite{sctd} (e.g., when the traffic flow on a segment of a road undergoes a sudden change, the traffic flows on adjacent segments also exhibit certain changes). Therefore, although the prediction results based on historical sensing data are inaccurate in some cases, they retain spatio-temporal information, making it possible to extract accurate implications among data points. Thus, our method has greater advantages than prediction-based methods, which directly use predictions.

\subsection{Reputation-Based TD Module}
TD methods can estimate the ground truth based on submitted data and the weights of their providers. Traditional TD methods require a large amount of data for the same target (e.g., multiple MUs that sense the traffic flow data of the same segment of a road at the same time) to accurately estimate the corresponding ground truth. However, in an MCS system, the distributions of MU locations and periods may be sparse, and the amount of sensing data available for each target may vary, resulting in inaccurate estimates when traditional TD methods are used. Therefore, we design a reputation-based TD module that considers data implications, which allows for more accurate identification of low-quality sensing data and malicious MUs in scenarios characterized by data sparsity. Before providing a detailed description of this module, we first present the definitions that are used in the module.

\begin{definition} \label{def-q}
	(Data quality level) The quality level of the sensing data point $d_{i,n}^t$ is denoted as $q(d_{i,n}^t)$ and satisfies $0 \leq q(d_{i,n}^t) \leq 1$. This quality level indicates the trustworthiness of the data point $d_{i,n}^t$. Additionally, the expected data quality level, denoted as $\hat{q}(d_{i,n}^t)\in \left[0,1\right]$, is an estimation of $q(d_{i,n}^t)$ when the actual $q(d_{i,n}^t)$ is unknown. The estimate is calculated based on the credibility probabilities (i.e., the reputations) of the MUs who provided similar or the same data points as $d_{i,n}^t$.
	
\end{definition}
Specifically, the closer the value of data point $d_{i,n}^t$ is to the ground truth, the higher the trustworthiness of this data point is, thereby making $q(d_{i,n}^t)$ closer to 1.

\begin{definition} \label{def-r}
	(Reputation) The reputation of MU $i$ at time slot $t$ is denoted as $r_i^t \in \left[0,1\right]$. Reputation $r_i^t$ represents the probability that MU $i$ is a normal and credible user. Furthermore, the reputation of an MU equals the mean data quality level of all the sensing data submitted by this MU \cite{tf}.
\end{definition}

\begin{figure}[!t]
	\centering{\includegraphics[width=0.85\columnwidth]{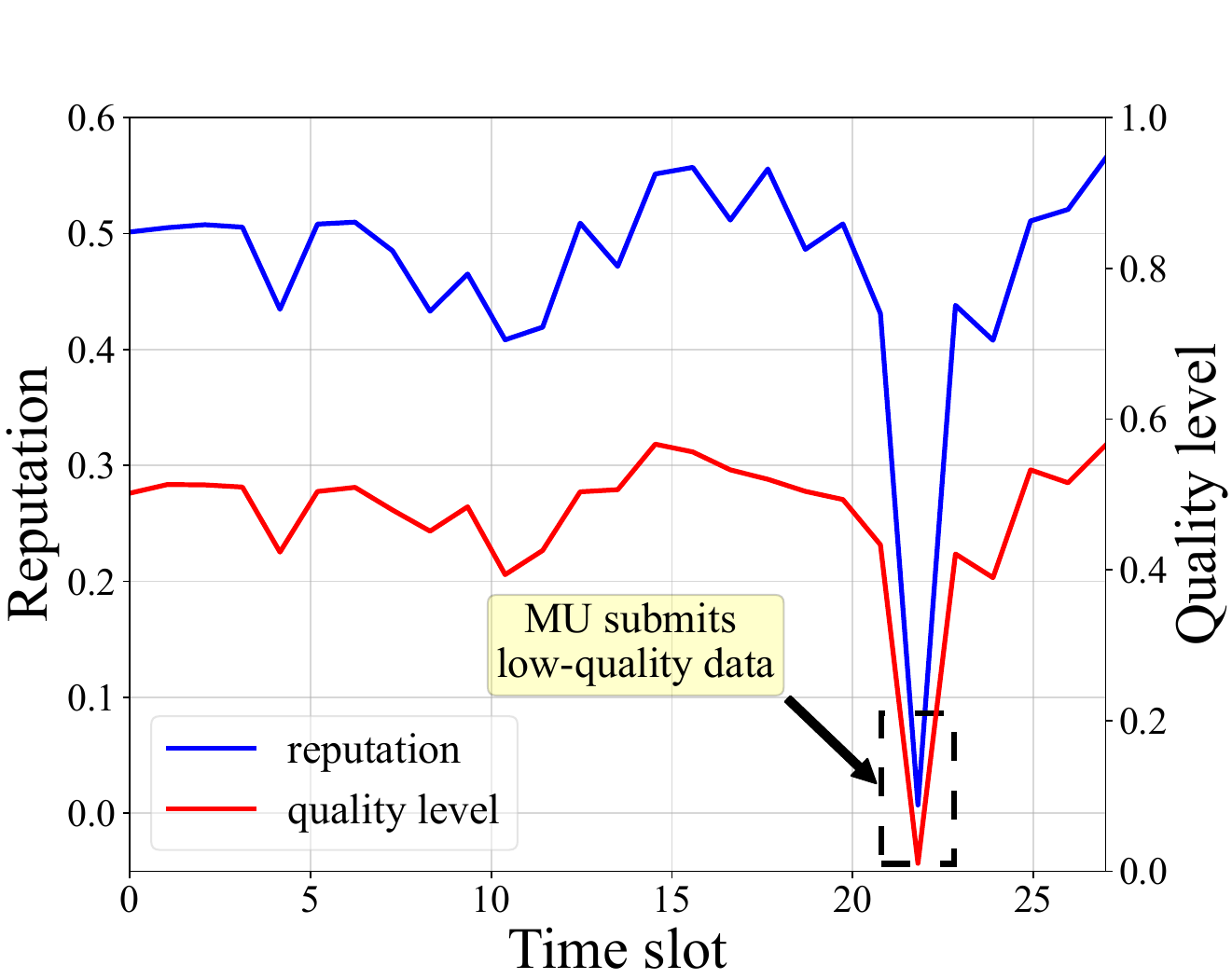}}
	\caption{Reputation of an MU influenced by data quality levels.}
	\label{fig-rep}
\end{figure}

Integrating a reputation mechanism into MCS can effectively reduce the misclassification of MUs. Even if normal MUs occasionally submit low-quality data under special circumstances (e.g., sudden device damage or abrupt weather changes), they can maintain a high reputation by consistently providing high-quality data over time \cite{reputation}. In contrast, malicious MUs, which consistently submit low-quality or fake data, will maintain a persistently low reputation. Fig. \ref{fig-rep} presents an example of how the reputation of a normal MU changes over time in relation to the quality levels of the submitted data. This example indicates that the reputations of MUs decrease when they submit low-quality sensing data but can remain high if they consistently submit high-quality data.

Like traditional TD methods \cite{mcstd, pptd, tf}, our reputation-based TD module consists of three steps, as follows.

\textbf{Calculate the quality levels of the sensing data.} We first consider the simple scenario shown in Fig. \ref{fig-scene}(a). This case includes two independent mobile users, MU $i$ and MU $j$, who collect data in region $n$ at time slot $t$. The values of their sensing data satisfy $v_{i,n}^t = v_{j,n}^t$. In this case, $v_{i,n}^t$ and $v_{j,n}^t$ are incorrect data values only if both MU $i$ and MU $j$ provide low-quality data simultaneously. Owing to the independence of the MUs, the probability of them providing simultaneous low-quality data should be $(1-r_i^t) (1-r_j^t)$. Therefore, the expected quality levels of data points $d_{i,n}^t$ and $d_{j,n}^t$ are given by $1-(1-r_i^t) (1-r_j^t)$, which represents the probability that data points $d_{i,n}^t$ and $d_{j,n}^t$ are correct. By generalizing the results to common scenarios, the expected quality level of data point $d_{i,n}^t$ can be calculated as follows:
\begin{equation}\label{eq-calq}
	\hat{q}(d_{i,n}^t) = 1-\!\prod_{i \in \bm{I}(v_{i,n}^t)}(1-r_i^t)
\end{equation}
where $\bm{I}(v_{i,n}^t)$ represents the set of all MUs that submit data with values equal to $v_{i,n}^t$ in region $n$ at time slot $t$.

However, $1-r_i^t$ may approach 0 when MU $i$ has a high reputation, leading to the disappearance of the cumulative product term in \eqref{eq-calq}. To address this problem, we define the reputation score of MU $i$ at time slot $t$ as
\begin{equation}\label{eq-rscore}
	R_i^t = -\ln(1-r_i^t)
\end{equation}
where $R_i^t \in \left[0,+\infty\right)$ as $r_i^t \in [0,1]$, and $\lim_{r_i^t \to 1}R_i^t=+\infty$. Similarly, the expected quality score of the sensing data point $d_{i,n}^t$ is defined as
\begin{equation}\label{eq-qscore}
	\widehat{Q}(d_{i,n}^t) = -\ln\left(1-\hat{q}(d_{i,n}^t)\right)
\end{equation}
where $\widehat{Q}(d_{i,n}^t) \in \left[0,+\infty\right)$ and $\lim_{\hat{q}(d_{i,n}^t) \to 1}\widehat{Q}(d_{i,n}^t)=+\infty$. Subsequently, we can rewrite \eqref{eq-calq} with the reputation score and the expected quality score as follows:
\begin{equation}\label{eq-calqscore}
	\widehat{Q}(d_{i,n}^t) = \sum_{i\in \bm{I}(v_{i,n}^t)}R_i^t.
\end{equation}

Moreover, the implications between sensing data points can influence the quality levels of these data. Taking the scenario shown in Fig. \ref{fig-scene}(a) as an example, the values of the data points submitted by MUs $i$ and $k$ satisfy $v_{i,n}^t \neq v_{k,n}^t$. When $imp(d_{i,n}^t,d_{k,n}^t) > 0$, data points $d_{i,n}^t$ and $d_{k,n}^t$ exhibit mutual support, so the trustworthiness of data point $d_{k,n}^t$ can increase the trustworthiness of $d_{i,n}^t$. Therefore, the quality level of data point $d_{i,n}^t$ increases the quality level of $d_{k,n}^t$. In contrast, the quality level of data point $d_{i,n}^t$ decreases the quality level of $d_{k,n}^t$ if $imp(d_{i,n}^t,d_{k,n}^t) < 0$. Hence, the overall data quality score needs to consider both the expected quality and the influence of the implication as follows:
\begin{equation}\label{eq-calQ}
	Q(d_{i,n}^t) = \widehat{Q}(d_{i,n}^t) + \rho\!\sum_{d_{j,n^{\prime}}^{t^\prime}\in \bm{D}^\prime}\!\widehat{Q}(d_{j,n^{\prime}}^{t^\prime})imp(d_{i,n}^t,d_{j,n^\prime}^{t^\prime})
\end{equation}
where $0<\rho<1$ is a hyperparameter that determines the proportion of the overall quality score influenced by the implication. $\bm{D}^\prime$ is a data cache that stores the sensing data derived from the previous $l$ time slots and is updated at the end of each time slot. When the hyperparameter $l$ is set to a larger value, more historical sensing data are considered when calculating the implications that influence the quality of the data. Thus, the data quality evaluation becomes more accurate, but the calculation time also increases. Finally, the quality level of the sensing data point $d_{i,n}^t$ can be calculated as follows:
\begin{equation}\label{eq-Q2q}
	q(d_{i,n}^t) = 1 - e^{-Q(d_{i,n}^t)}.
\end{equation}

\textbf{Calculate the reputations of the MUs.} The reputations of all the MUs need to be updated at each time slot $t \in \left\{1,2,...,T\right\}$. First, we initialize the reputation for every MU at time slot $t$ as
\begin{equation}\label{eq_init_rep}
	r_i^t = r_i^{t-1}, \forall i \in \left\{1,2,...,N\right\}.
\end{equation}
Note that when $t=1$, $r_i^t$ is initialized as $r_i^0$, which is the initial reputation of MU $i$ set by the sensing platform. Since each MU submits data at most once per time slot, we need only update the reputations of the MUs who submit data at the current time slot $t$. Specifically, the reputations of these MUs are updated based on the quality of the data they submit at time slot $t$. The reputation update function is designed according to \cite{design_rep}:
\begin{equation} \label{eq-updater}
	r_i^t = \begin{cases}
		\begin{aligned}
			\frac{1}{2} \bigg( &1 +\alpha \Big( \frac{q(d_{i,n}^t) -\gamma}{1-\gamma} \Big) \\
			&+ \left(1 - \alpha \right) \left(2r_i^t - 1 \right) \bigg), 
		\end{aligned} & \gamma \leq q(d_{i,n}^t) \leq 1 \\
		\begin{aligned}
			\frac{1}{2} \bigg( &1 +\beta \Big( \frac{q(d_{i,n}^t) -\gamma}{\gamma} \Big) \\
			&+ \left(1 - \beta \right) \left(2r_i^t - 1 \right) \bigg), 
		\end{aligned} & 0 \leq q(d_{i,n}^t) < \gamma. \\
	\end{cases}
\end{equation}
where $\alpha$ and $\beta$ are hyperparameters that determine the increasing and decreasing rates of the reputation update function, which satisfies $0<\alpha<\beta<1$, and $\gamma$ is the quality threshold set by the platform, which satisfies $0<\gamma<1$. Since the data quality level represents the trustworthiness of the data, sensing data with quality levels that are greater than or equal to $\gamma$ are identified as high-quality data, whereas data with quality levels lower than $\gamma$ are considered low-quality data. According to \eqref{eq-updater}, the reputations of MUs who have submitted high-quality data increase, whereas those of MUs who have submitted low-quality data decrease.

The reputation update function shown in \eqref{eq-updater} satisfies the demands mentioned in \cite{design_rep} and \cite{reputation} for monitoring the reputations of users in long-term multi-user scenarios and thus can be applied in MCS systems.

\begin{figure}[!t]
	\centering{\includegraphics[width=0.9\columnwidth]{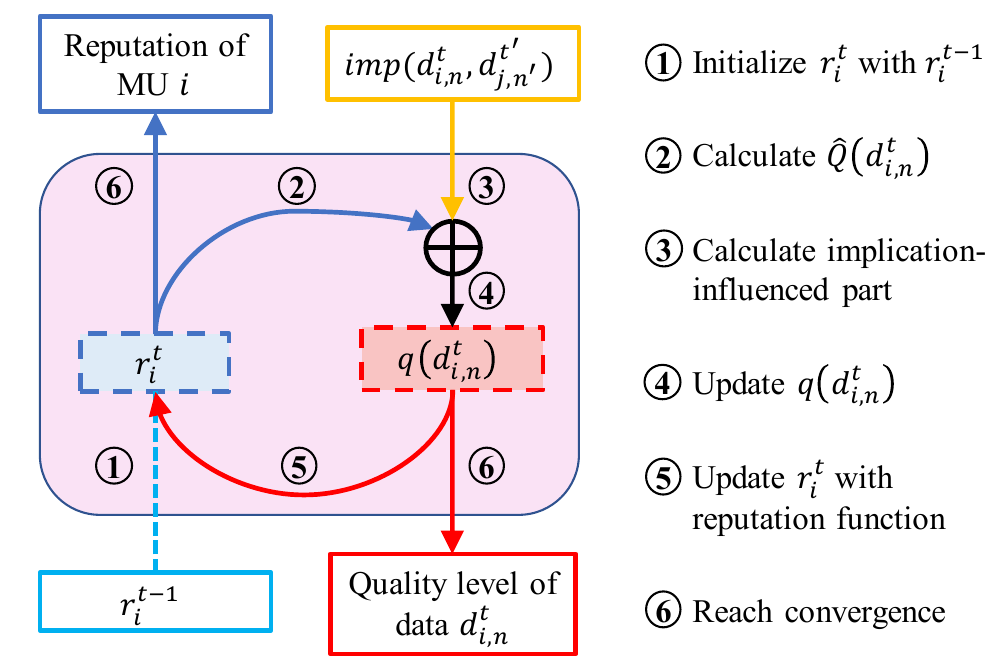}}
	\caption{Workflow of the reputation-based TD module.}
	\label{fig-td}
\end{figure}

\textbf{Update the data quality levels and the reputations of the MUs alternately until convergence is reached}. When the reputations of the MUs are updated according to \eqref{eq-updater}, the expected quality levels of the data submitted by these MUs may change accordingly. Consequently, the data quality levels should be updated again according to \eqref{eq-calqscore} to \eqref{eq-Q2q}. Moreover, since the data quality level $q(d_{i,n}^t)$ changes, the reputation of MU $i$ needs to be recalculated by \eqref{eq-updater}. Therefore, as shown in Fig. \ref{fig-td}, the data quality level $q(d_{i,n}^t)$ and the MU reputation $r_i^t$ are alternately updated until convergence is achieved, resulting in the final quality level of data point $d_{i,n}^t$ and the final reputation of MU $i$ at time slot $t$.

\begin{algorithm}[!t]
	\SetAlgoLined
	\DontPrintSemicolon
	\KwIn{Sensing data $\bm{D}^t = \left\{d_{i,n}^{t}\right\}_{i=1,n=1}^{I,N}$ and MU reputations $\left\{r_i^{t-1}\right\}_{i=1}^I$}
	\vspace{1pt}
	\KwOut{Data quality levels $\left\{q(d_{i,n}^{t})\right\}_{i=1,n=1}^{I,N}$, MU reputations $\left\{r_i^t\right\}_{i=1}^I$, and high-quality data cache $\bm{\widehat{D}}^t$}
	\vspace{1pt}
	Initialize $\bm{\widehat{D}}^t \gets \varnothing$, $r_i^{t} \gets r_i^{t-1}$ for $i = 1,2,...,I$\\
	Predict $\widehat{\bm{G}}^t$ with the CFSTTN\\
	\If{$t > l$}{
		Remove old data from the cache: $\bm{D}^\prime \gets \bm{D}^\prime \setminus \bm{D}^{t-l}$\
	}
	Add new data to the cache:
	$\bm{D}^\prime \gets \bm{D}^\prime \cup \bm{D}^{t}$\\
	\Repeat{$\varDelta_{max} < \epsilon$}{
		\For{$i=1,2,...,I$}{
			$r_i^{old} \gets r_i^{t}$\\
			$\varDelta_i \gets 0$
		}
		\For{$n = 1,2,...,N$}{
			Calculate $c_n^t$ according to \eqref{eq-cnt}\\
			Calculate $\dot{\delta}_n^{t}$ with $c_n^t$ according to \eqref{eq-cre}\\
		}
		\For{$d_{i,n}^t \in \bm{D}^t$}{
			\For{$d_{j,n^\prime}^{t^\prime} \in \bm{D}^\prime$}{
				Calculate $imp(d_{i,n}^t,d_{j,n^\prime}^{t^\prime})$ according to \eqref{eq13}\\
			}
			Calculate $q(d_{i,n}^t)$ according to \eqref{eq-calq} to \eqref{eq-Q2q}\\
			Update $r_i^t$ with $q(d_{i,n}^t)$ according to \eqref{eq-updater}\\
			Calculate $\varDelta_i = \left|r_i^t - r_i^{old}\right|$\\
		}
		Calculate $\varDelta_{max} = \max \left\{\varDelta_i\right\}_{i=1}^I$\\
	}
	\For{$d_{i,n}^t \in \bm{D}^t$}{
		\If{$q(d_{i,n}^t) \geq \gamma$}{
			$\bm{\widehat{D}}^t \gets \bm{\widehat{D}}^t \cup d_{i,n}^t$\
		}
	}
	\caption{Enhancing Data Quality with PRBTD}
	\label{alg1}
\end{algorithm}

\subsection{Data Enhancement with PRBTD}
Finally, we present Algorithm \ref{alg1}, which outlines the process of enhancing the quality of the sensing data with the proposed PRBTD method at time slot $t$. First, as shown in Fig. \ref{fig-prbtd}, the PRBTD method predicts the ground truth $\widehat{\bm{G}}^t$ via the CFSTTN prediction module. The data feature and implication calculation module subsequently uses the prediction results to calculate the sensing errors of the submitted data as features and uses these features to determine the degrees of implication between the data. The PRBTD method ultimately evaluates the quality levels of the data and the reputations of the MUs with the reputation-based TD module while considering the implications between the data. Moreover, sensing data with quality levels lower than $\gamma$ are eliminated, whereas the remaining data are stored in a high-quality data cache $\bm{\widehat{D}}^t$, thus achieving data cleansing and quality enhancement. Similarly, PRBTD can identify malicious MUs in MCS systems. After the task begins, the sensing platform can periodically check the reputations of the MUs and identify those whose reputations consistently remain below the predefined threshold as malicious MUs.

\section{Experiments and Analysis}
\subsection{Experimental Setup}
\subsubsection{Dataset}
The TaxiBJ dataset \cite{stresnet} is utilized for conducting simulation experiments in our MCS system to validate the ability of PRBTD to enhance the quality of the sensing data. This dataset comprises real-world taxi flow data with spatio-temporal correlations derived from Beijing city, making it suitable for simulating MCS systems. Specifically, the TaxiBJ dataset contains 7220 entries of data spanning four time intervals: 1st Jul. 2013 -- 30th Oct. 2013, 1st Mar. 2014 -- 30th Jun. 2014, 1st Mar. 2015 -- 30th Jun. 2015, and 1st Nov. 2015 -- 10th Apr. 2016. The data were sampled at 30-minute intervals within each period. In addition, the dataset partitions Beijing into a grid of $32\times32$ spatial regions based on longitude and latitude information.

\subsubsection{Setup}
In our simulation, the sensing task spans $T=120$ continuous time slots, each lasting 30 minutes, and covers $N=32$ regions. Specifically, we use the data from the last 120 time slots within 32 randomly selected regions of the TaxiBJ dataset for the simulations. These data are regarded as the ground truth of the sensing data in the task. Our MCS system includes 90 normal MUs and 10 malicious MUs that participate in the sensing task. We set the sensing errors of normal MUs to 0, ensuring that the sensing data they submit are equal to the corresponding ground truth. However, the sensing errors of malicious MUs are sampled from a Gaussian distribution $N(\mu, \sigma^2)$. Specifically, we set $\mu=0.3$ and $\sigma=0.15$ as the basic setup to simulate the sensing errors carried by the data provided by malicious MUs. Additionally, each MU is set to submit sensing data a total of $k$ times during the sensing task, where $0<k<T$. In our experiments, we set $k=30$ to simulate a scenario where MUs do not need to collect and submit data frequently. When calculating the reputations of MUs, we initialize the reputation of each MU as $r_i^0=0.5$ and set the data quality threshold as $\gamma=0.5$. The reason is that the reputations of MUs and the quality levels of data represent their respective probabilities of trustworthiness, making 0.5 a suitable initial value and threshold. Moreover, the increasing and decreasing rates in the reputation update function \eqref{eq-updater} are set as $\alpha=0.018$ and $\beta=0.06$, respectively, whereas the proportion of the data quality levels influenced by implications is set as $\rho=0.02$. These parameters are hyperparameters in our method, and their specific settings were determined through a series of preliminary experiments. Finally, the detailed parameter settings for the simulations are listed in Table \ref{table-parameter}.

\begin{table}[!t]
	\caption{Simulation Parameters}
	\label{table-parameter}
	\centering
	\setlength{\tabcolsep}{3pt}
	\renewcommand{\arraystretch}{1.3}
	\begin{tabular}{p{160pt}|p{60pt}}
		\hline
		Parameters&
		Values\\
		\hline
		Number of MUs ($I$)&
		100\\
		Number of times a MU performs the task ($k$)&
		30\\
		Initial reputation of each MU ($r_i^0$)&
		0.5\\
		Mean of the sensing errors induced by malicious MUs ($\mu$)&
		0.3\\
		Standard deviation of the sensing errors induced by malicious MUs ($\sigma$)&
		0.1\\
		Increase rate of reputation ($\alpha$)&
		0.018\\
		Decrease rate of reputation ($\beta$)&
		0.06\\
		The data quality threshold ($\gamma$)&
		0.5\\
		Proportion of the implication-influenced part of the quality score ($\rho$)&
		0.02\\
		Size of the data cache $\bm{D}^\prime$ ($l$)&
		5\\
		Number of time slots used as the prediction input ($p$)&
		5\\
		Ending factor of PRBTD ($\epsilon$)&
		0.001\\
		\hline
	\end{tabular}
\end{table}

In the CFSTTN, we set the network parameters in accordance with the setup used by the GLSTTN \cite{glsttn}. The first 7100 time slots of data derived from the TaxiBJ dataset are used for training, whereas the last 120 time slots are used for validation. We adopt the adaptive moment estimation (Adam) optimizer with an initial learning rate of 0.001 for training. This learning rate is reduced to $10^{-4}$ at 50\% of the total training epochs and further reduced to $10^{-5}$ at 75\% of the total. The total number of training epochs is set to 300, with a batch size of 32 per epoch. Additionally, we set the prediction input of the CFSTTN module to consist of data from the preceding 3 time slots along with data from 1 day and 1 week prior. We implement the CFSTTN via PyTorch 1.1.0. All the experiments are conducted on a system equipped with an NVIDIA 1080Ti GPU. Our source code is publicly available at \url{https://github.com/glab2019/PRBTD}.

\subsection{Evaluation of the CFSTTN}
\begin{figure}[!t]
	\centering{\includegraphics[width=0.9\columnwidth]{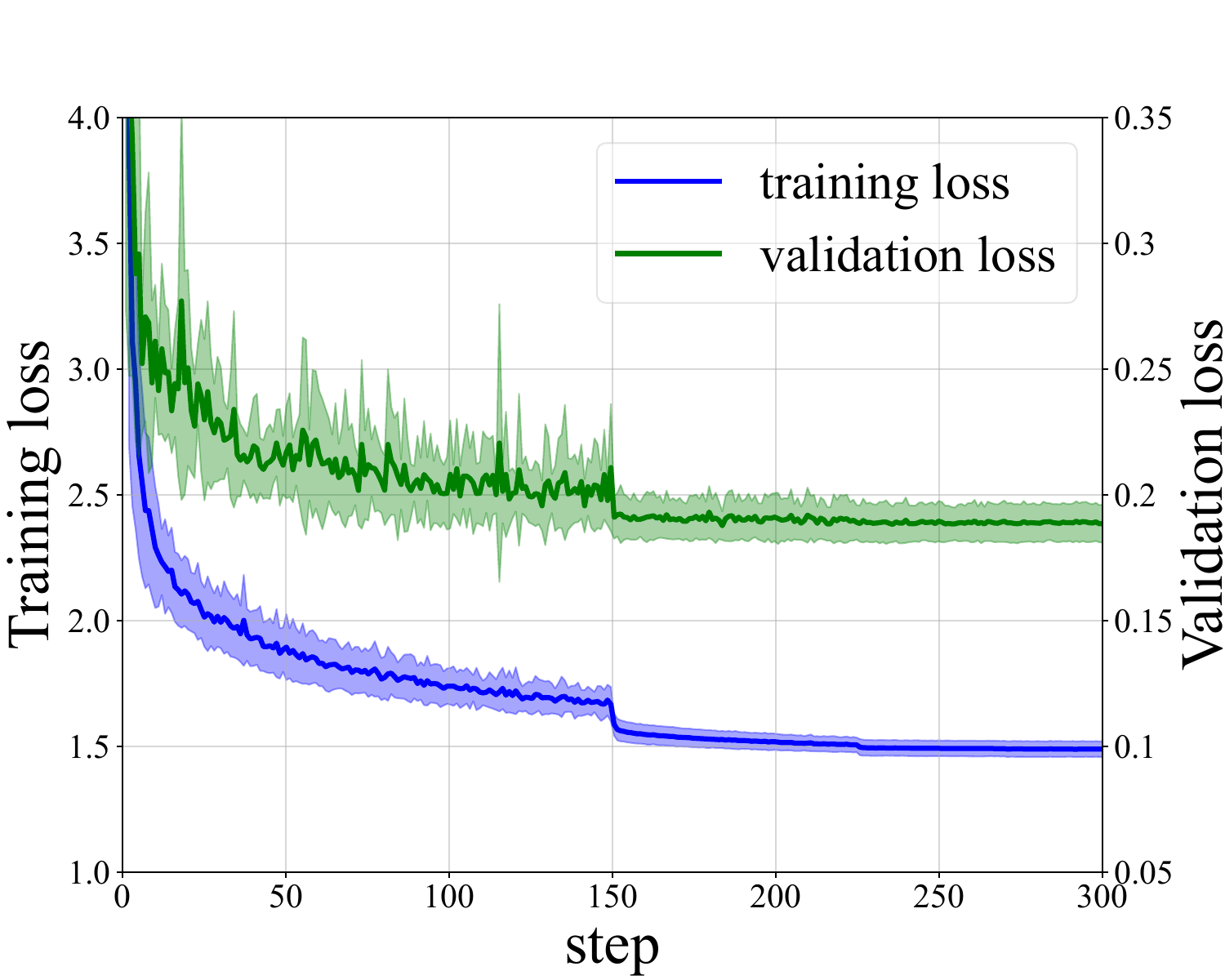}}
	\caption{Convergence of the CFSTTN.}
	\label{fig-loss}
\end{figure}

We first evaluate the convergence performance of the CFSTTN prediction module in our PRBTD framework. The training and validation loss curves produced by the CFSTTN on the TaxiBJ dataset are shown in Fig. \ref{fig-loss}. The CFSTTN converges after approximately 220 epochs, with an average training time of 82.56 seconds per epoch, leading to an average convergence duration of 5.04 hours. These results were obtained after repeating the training 6 times and indicate that the convergence time of the CFSTTN is acceptable.

We analyze the prediction performance of the CFSTTN via the root mean square error (RMSE) and the coefficient of determination (R2). The RMSE measures the difference between the prediction results and the ground truth. Therefore, a lower RMSE reflects better performance. R2 represents the explanatory ability of the prediction method. An R2 closer to 1 indicates a higher prediction accuracy. In our experiment, which used the TaxiBJ dataset, the CFSTTN achieves an RMSE of 16.88 and an R2 of 0.9583. Moreover, we compare our CFSTTN with the following 6 baselines:
\begin{itemize}
	\item \textbf{ARIMA.} The auto-regressive integrated moving average (ARIMA) is a popular time series prediction model that combines auto-regression, differencing, and moving average to predict future values.
	\item \textbf{SARIMA.} The seasonal auto-regressive integrated moving average (SARIMA) extends ARIMA by including seasonal components, allowing it to model and predict data with seasonal patterns.
	\item \textbf{VAR.} The vector auto-regressive (VAR) model is a multivariate time series model that captures linear interdependencies among multiple time series.
	\item \textbf{ST-ANN.} The spatio-temporal artificial neural network (ST-ANN) extracts spatial and temporal features and uses an artificial neural network to model complex dependencies for prediction.
	\item \textbf{DeepST.} This deep neural network-based model employs convolutional neural networks (CNNs) to model spatial dependencies and recurrent neural networks (RNNs) to capture temporal dependencies.
	\item \textbf{ST-ResNet.} The spatio-temporal residual networks (ST-ResNet) utilize residual learning to capture spatial and temporal dependencies in data.
\end{itemize}
The model comparisons are presented in Table \ref{table-rmse}, demonstrating that our CFSTTN outperforms the baselines.

\begin{table}[!t]
	\caption{Comparison Among the RMSEs Induced by Different Methods on TaxiBJ}
	\label{table-rmse}
	\centering
	\setlength{\tabcolsep}{10pt}
	\renewcommand{\arraystretch}{1.2}
	\begin{tabular}{p{90pt}|p{40pt}}
		\hline
		Method&
		RMSE\\
		\hline
		ARIMA&
		22.78\\
		SARIMA&
		26.88\\
		VAR&
		22.88\\
		ST-ANN&
		19.57\\
		DeepST&
		18.18\\
		ST-ResNet&
		17.30\\
		\textbf{CFSTTN (ours)}&
		\textbf{16.88}\\
		\hline
	\end{tabular}
\end{table}

\begin{figure*}[!t]
	\hspace{2mm}
	\centering
	\begin{minipage}[b]{0.31\linewidth}
		\includegraphics[width=\linewidth]{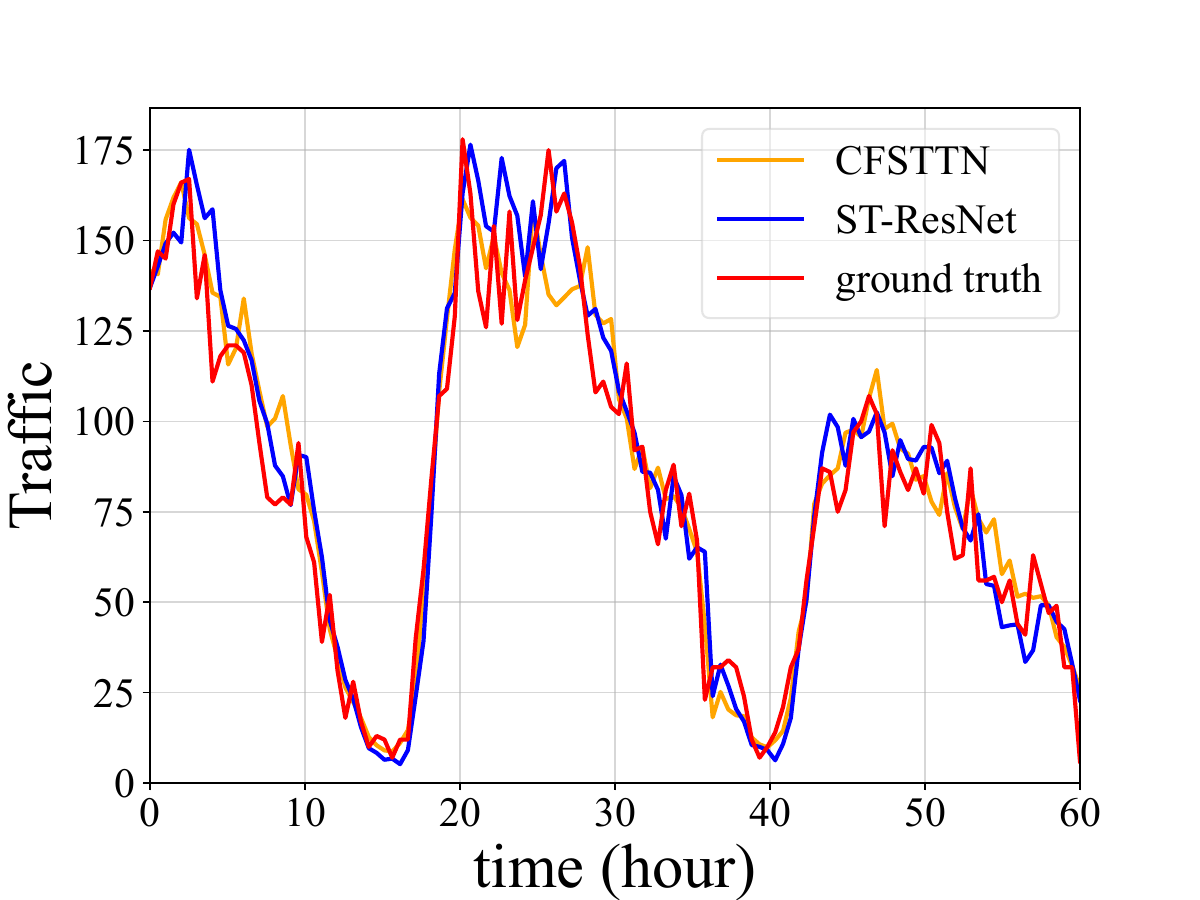}
		\centering {\fontsize{8pt}{\baselineskip}\selectfont (a) Ground-truth and prediction curves.}
		\label{fig-prediction-a}
	\end{minipage}
	\hspace{2.5mm}
	\begin{minipage}[b]{0.31\linewidth}
		\includegraphics[width=\linewidth]{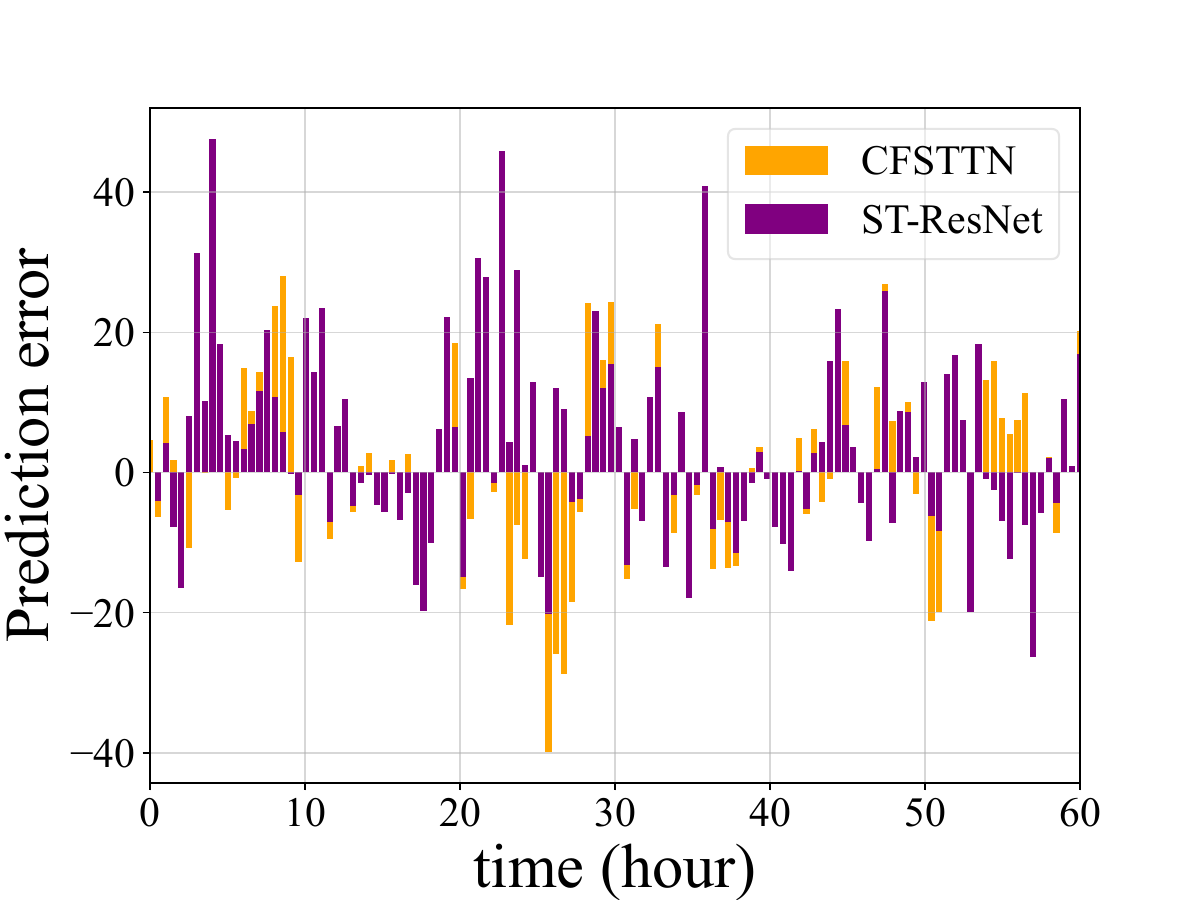}
		\par\hspace{5pt}\centering {\fontsize{8pt}{\baselineskip}\selectfont (b) Prediction errors.}
		\label{fig-prediction-b}
	\end{minipage}
	\hspace{2.5mm}
	\begin{minipage}[b]{0.31\linewidth}
		\includegraphics[width=\linewidth]{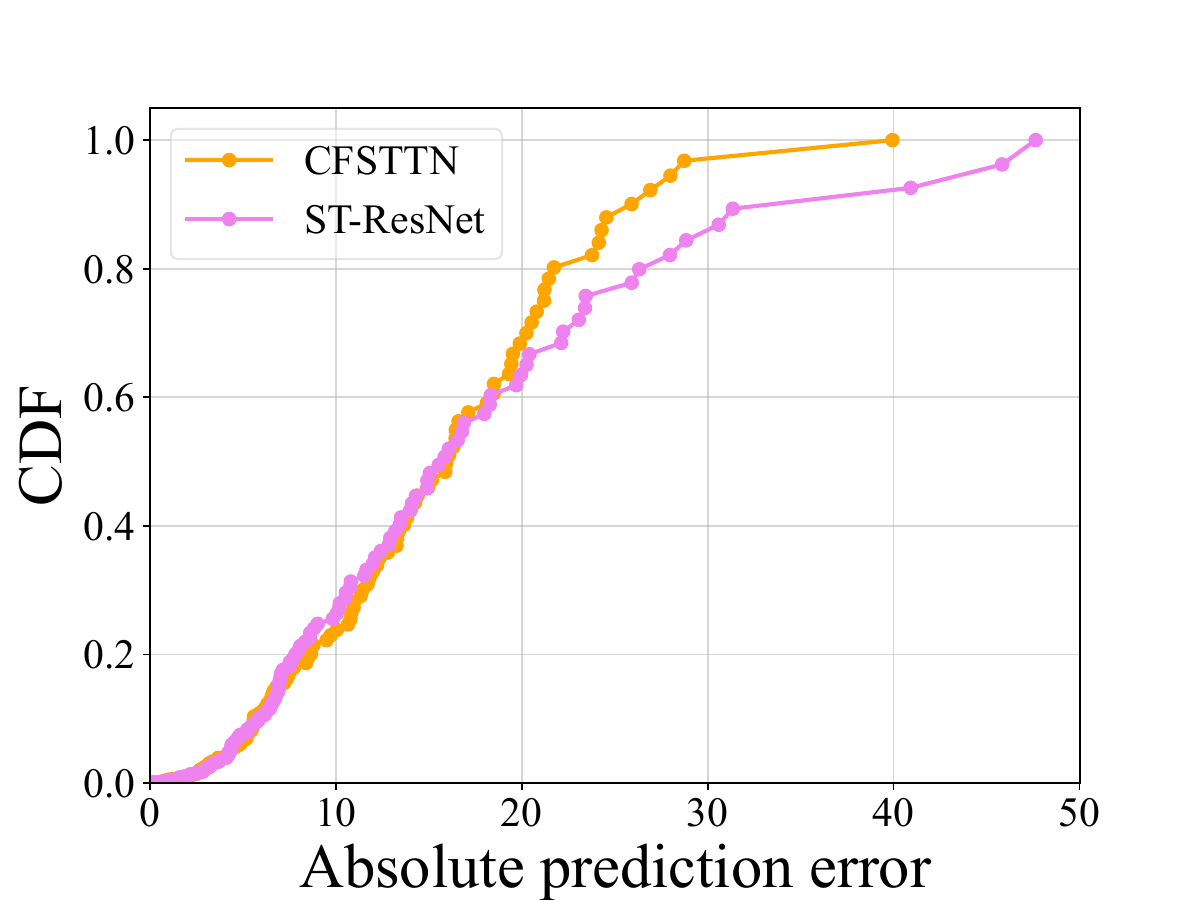}
		\par\hspace{5pt}\centering {\fontsize{8pt}{\baselineskip}\selectfont (c) CDF of the absolute prediction error.}
		\label{fig-prediction-c}
	\end{minipage}
	\caption{Comparison of prediction results for region (24,9).}
	\label{fig-prediction}
\end{figure*}

Finally, we analyze the ability of the CFSTTN by visualizing its prediction results obtained on the TaxiBJ dataset and comparing them with the predictions of ST-ResNet, which is the best-performing method among the baselines. We take region (24,9) as an example. Fig. \ref{fig-prediction}(a) shows that the prediction results of the CFSTTN for this region closely align with the ground truth and are better than those of ST-ResNet. Moreover, Figs. \ref{fig-prediction}(b) and (c) show that the errors of the CFSTTN in the prediction results are smaller than those of ST-ResNet, thus highlighting that the CFSTTN has a good prediction ability.

\subsection{Experimental Results}
We analyze the performance of the PRBTD method based on the following metrics.

\begin{itemize}
	\item \textbf{F1-score.} F1-scores of the methods when classifying normal and malicious MUs. The F1-score is a metric that combines precision and recall and can be used to evaluate the ability of a method to correctly classify normal and malicious MUs in the MCS system.
	\item \textbf{Reputation distance.} Interclass reputation distance between normal MUs and malicious MUs. A larger reputation distance indicates better performance in terms of classifying normal MUs and malicious MUs in MCS systems.
	\item \textbf{Noise reduction ratio.} Proportion of noise removed relative to the total noise after applying the enhancement method. Specifically, we calculate the noise reduction ratio as 1 minus the ratio of the average noise in the enhanced data to the average noise in the original data. A higher noise reduction ratio indicates better performance in identifying low-quality data and enhancing data quality.
\end{itemize}

Each set of experiments in our simulations is repeated 6 times, and the results reflect the mean results obtained across the repetitions.

\subsubsection{Baseline Comparisons}
We compare PRBTD with the following 4 baseline methods.
\begin{itemize}
	\item \textbf{Weight truth inference (WEI).} A data aggregation method. This method calculates the average of all the data in region $n$ prior to time slot $t$ as the ground truth.
	\item \textbf{CFSTTN-based (CNB).} A prediction-based method. This method directly uses the prediction results obtained from the CFSTTN as the ground truth.
	\item \textbf{TD \cite{generaltf}.} A traditional user-weight-based TD method. This method iteratively estimates the ground truth based on the data submitted by MUs and the weights of these MUs.
	\item \textbf{Decentralized trust inference (DTI) \cite{decentralized}.} A data filling method. This method uses trusted baseline data (e.g., data collected by dispatching UAVs during the task) to construct a sparse data matrix, which is then completed via Bayesian probabilistic matrix factorization (BPMF). Specifically, we utilize historical data as trusted baseline data when deploying DTI in our experiments. The completed matrix serves as the estimated ground truth.
\end{itemize}

\begin{table}[!t]
	\caption{Performances of the PRBTD and Baseline Methods in the Scenario with the Basic Setup}
	\label{table-baseline}
	\centering
	\setlength{\tabcolsep}{3pt}
	\renewcommand{\arraystretch}{1.3}
	\begin{tabular}{p{70pt}|p{45pt}|p{45pt}|p{58pt}}
		\hline
		Method& F1-score & Reputation distance & Noise reduction ratio\\
		\hline
		WEI & 0.9667 & 0.1004 & 0.2687\\
		CNB & 0.9741 & 0.1420 & 0.2863\\
		TD & 0.9778 & 0.2951 & 0.4252\\
		DTI & 0.9870 & 0.1562 & 0.3926\\
		\textbf{PRBTD (ours)} & \textbf{0.9907} & \textbf{0.3871} & \textbf{0.8100}\\
		\hline
	\end{tabular}
\end{table}

We choose these methods as baselines due to their reproducibility, strong representativeness and high performance among various types of methods, as well as their good adaptability to our MCS scenario. To use these methods in our experiments, we adjusted them to evaluate the quality of a sensing data point based on the distance between the data value and the estimated ground truth and update the reputations of the MUs according to \eqref{eq-updater}.

We first analyze the runtime efficiency of our PRBTD method. Specifically, in our experiments, the average runtime of the PRBTD method per slot was 0.5097 seconds, which is significantly less than the 30-minute duration we set for each time slot. Therefore, in our experiments, the PRBTD method is able to perform real-time data quality analysis.

The performances of the PRBTD method and baselines are presented in Table \ref{table-baseline}, which shows that PRBTD outperforms the other methods in terms of the three metrics. Specifically, the proposed PRBTD framework achieves the highest F1-score and a 31\% improvement in reputation distance over TD, the most commonly used method in MCS. Additionally, the noise reduction ratio of PRBTD is 0.8100, meaning that after executing the data quality enhancement via PRBTD, the average noise in the sensing data decreases by 81\%. This reduction demonstrates the strong denoising efficiency of PRBTD over the baselines.

\begin{figure*}[!t]
	\par\hspace{6.5pt}\centering
	\begin{minipage}[b]{0.31\linewidth}
		\includegraphics[width=\linewidth]{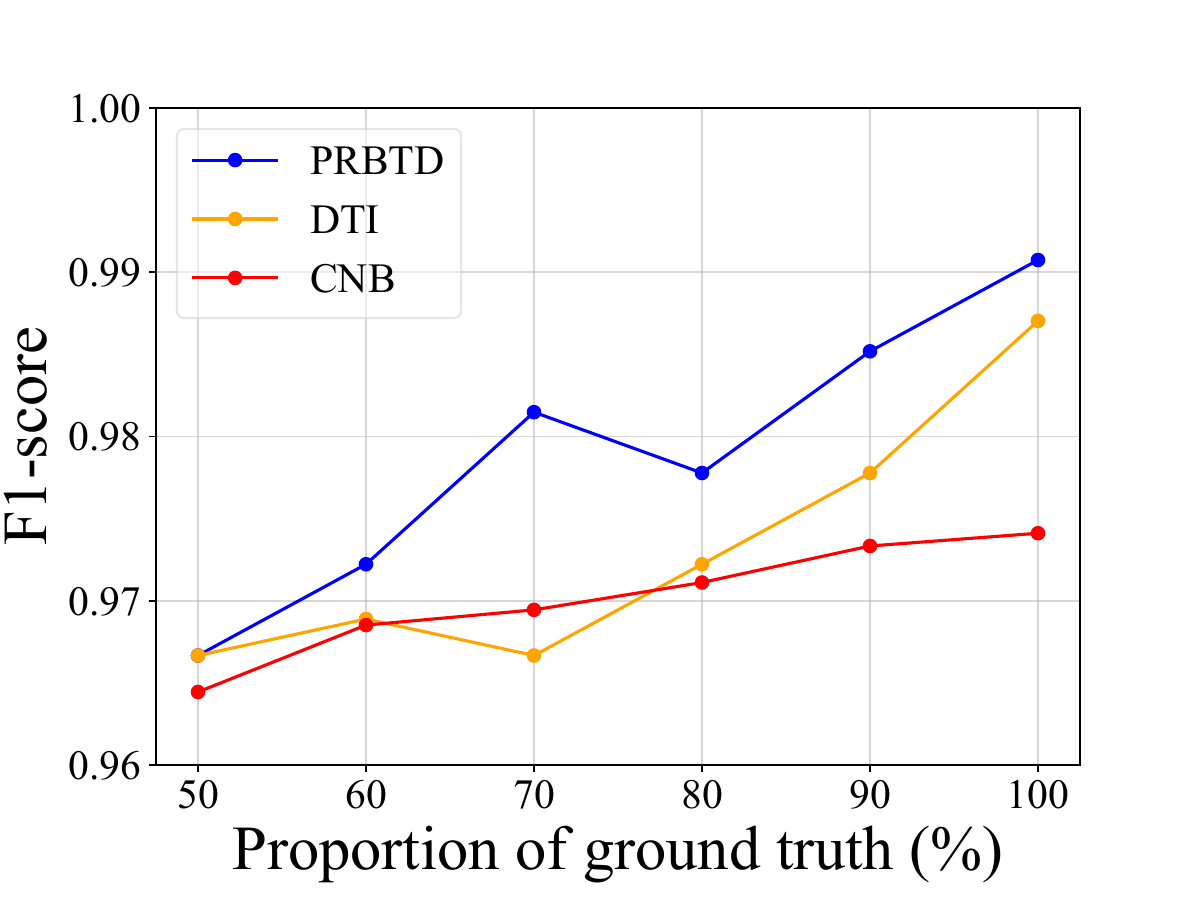}
		\centering {\fontsize{8pt}{\baselineskip}\selectfont (a) F1-score.}
		\label{fig-proportion-a}
	\end{minipage}
	\hspace{3.0mm}
	\begin{minipage}[b]{0.31\linewidth}
		\includegraphics[width=\linewidth]{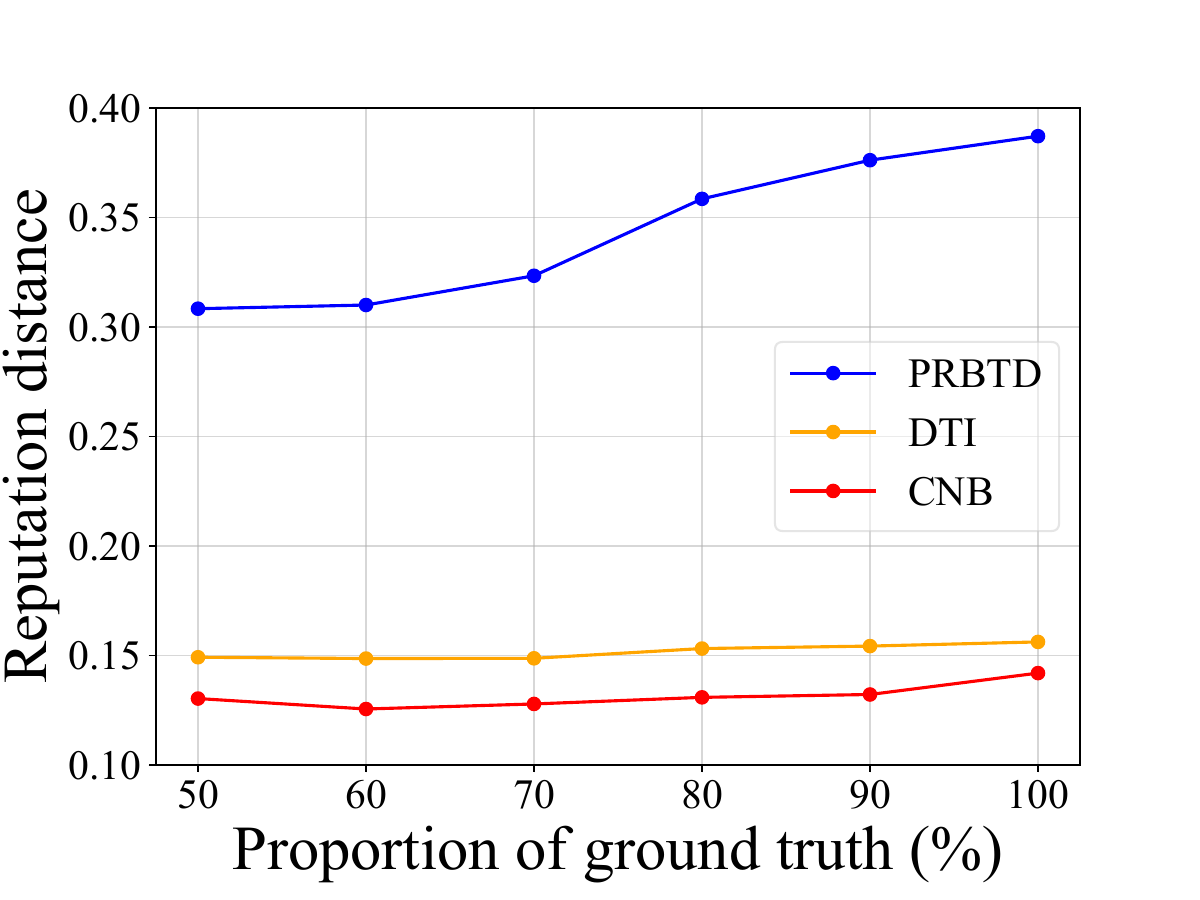}
		\centering {\fontsize{8pt}{\baselineskip}\selectfont (b) Reputation distance.}
		\label{fig-proportion-b}
	\end{minipage}
	\hspace{3.0mm}
	\begin{minipage}[b]{0.31\linewidth}
		\includegraphics[width=\linewidth]{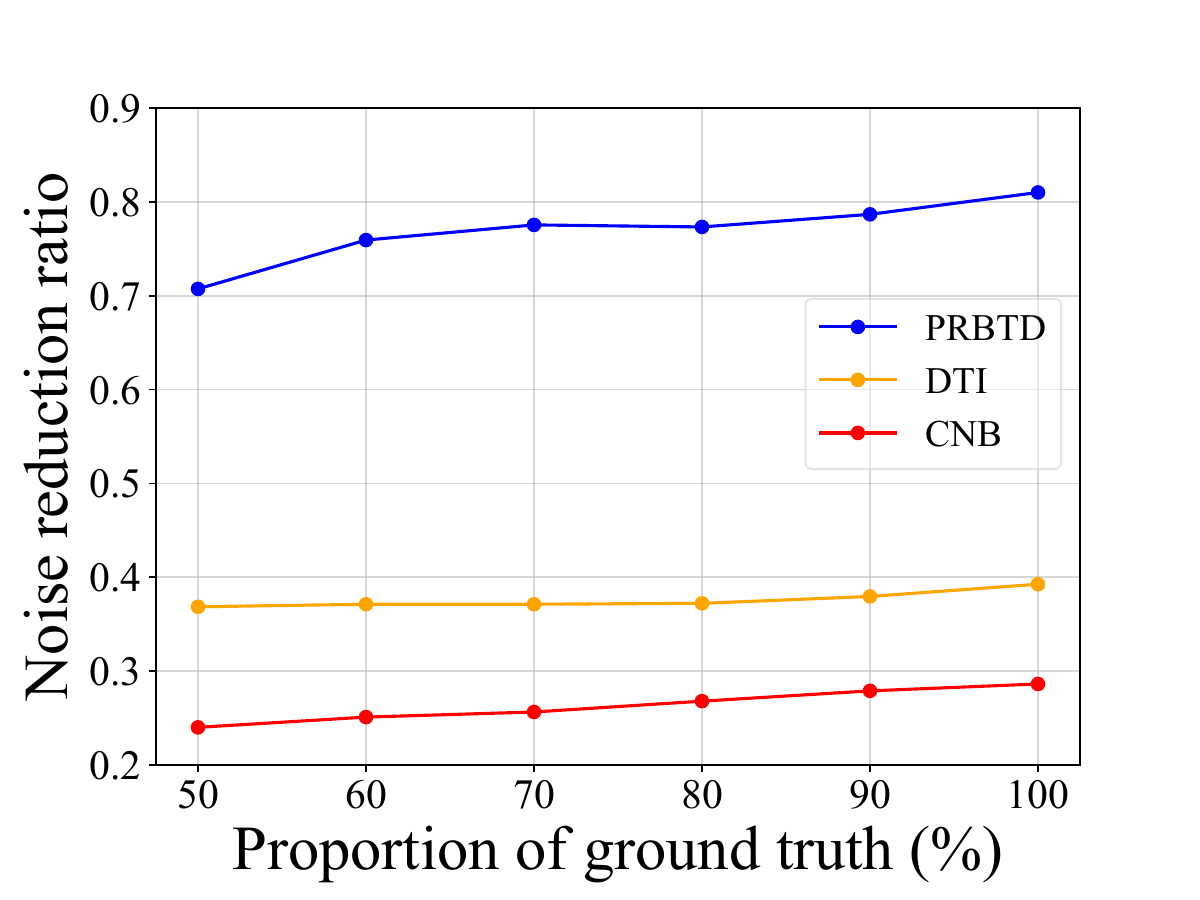}
		\centering {\fontsize{8pt}{\baselineskip}\selectfont (c) Noise reduction ratio.}
		\label{fig-proportion-c}
	\end{minipage}
	\caption{Comparisons among the PRBTD and baseline methods with different proportions of ground truth data used as historical data.}
	\label{fig-proportion}
\end{figure*}

\begin{figure*}[!t]
	\centering
	\begin{minipage}[b]{0.30\linewidth}
		\includegraphics[width=\linewidth]{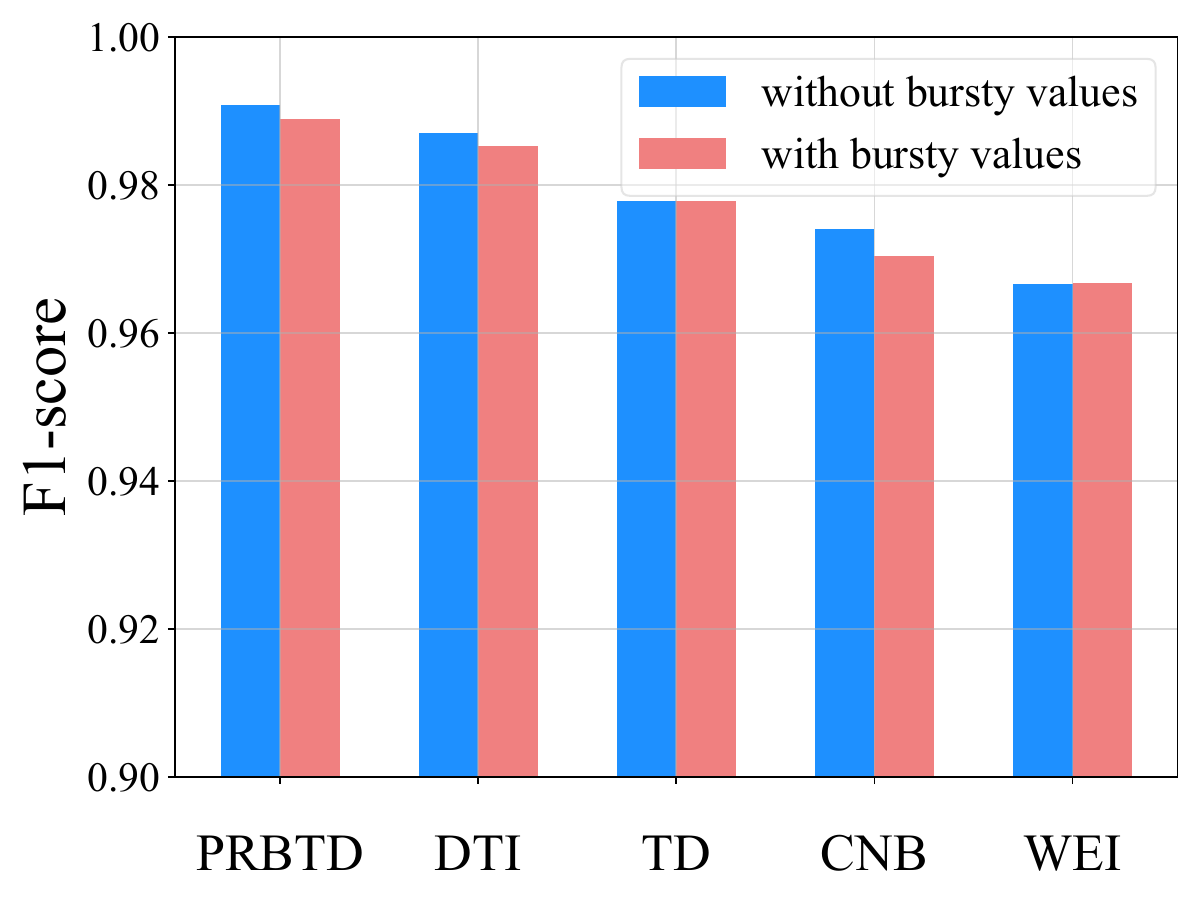}
		\par\hspace{14pt}\centering {\fontsize{8pt}{\baselineskip}\selectfont (a) F1-score.}
		\label{fig-bursty-a}
	\end{minipage}
	\hspace{5mm}
	\begin{minipage}[b]{0.30\linewidth}
		\includegraphics[width=\linewidth]{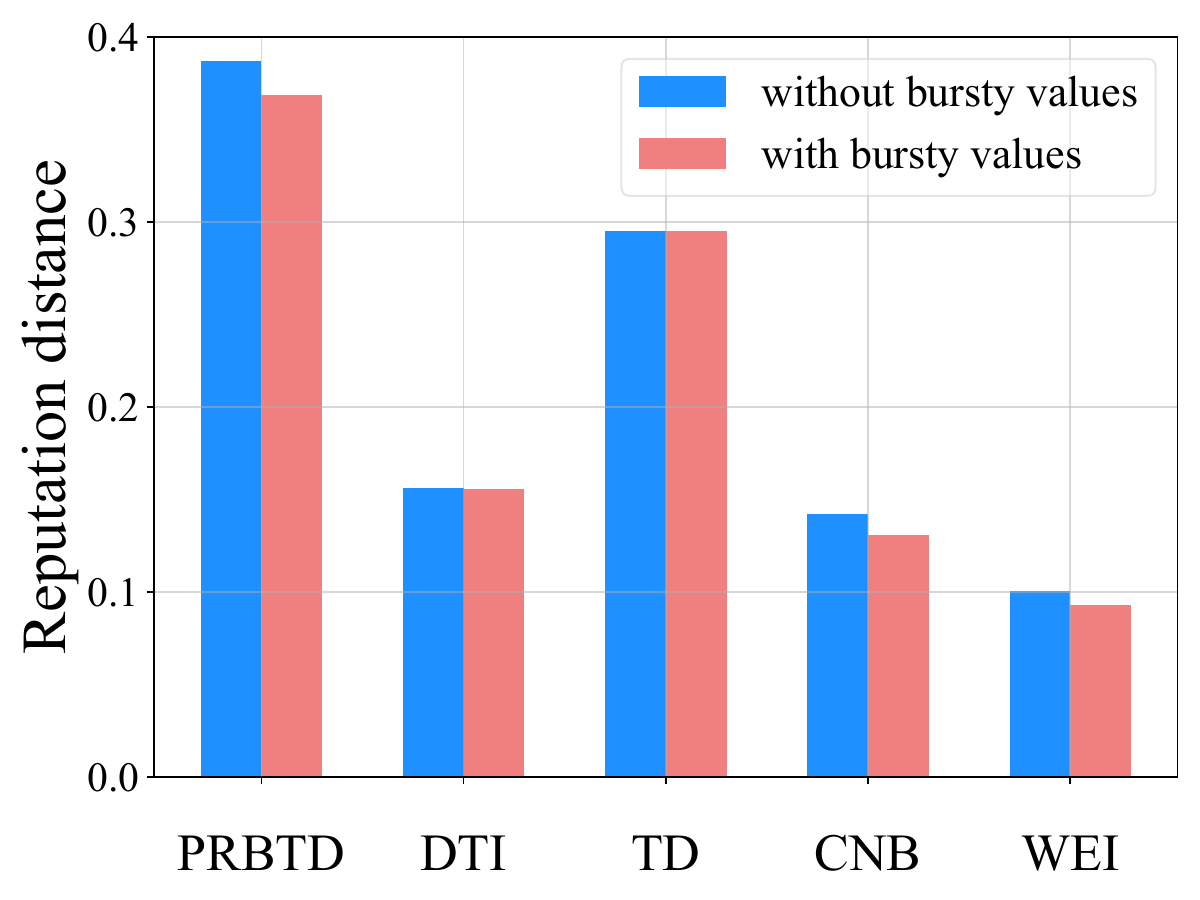}
		\par\hspace{14pt}\centering {\fontsize{8pt}{\baselineskip}\selectfont (b) Reputation distance.}
		\label{fig-bursty-b}
	\end{minipage}
	\hspace{5mm}
	\begin{minipage}[b]{0.30\linewidth}
		\includegraphics[width=\linewidth]{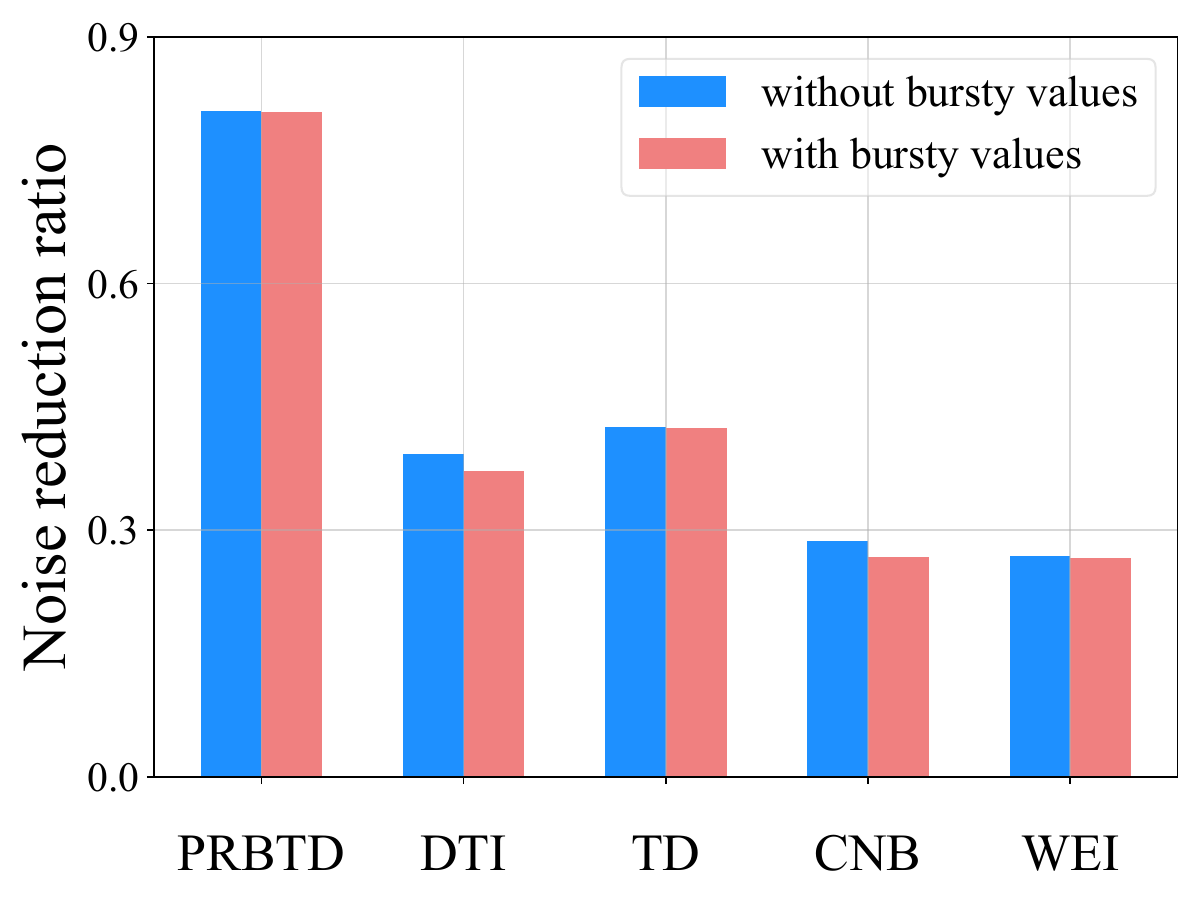}
		\par\hspace{12pt}\centering {\fontsize{8pt}{\baselineskip}\selectfont (c) Noise reduction ratio.}
		\label{fig-bursty-c}
	\end{minipage}
	\caption{Comparisons among the PRBTD and baseline methods in scenarios with and without bursty data values.}
	\label{fig-bursty}
\end{figure*}

\subsubsection{Experiments in Scenarios with Noisy Historical Data}
In practical MCS scenarios, historical data usually contain a certain amount of noise, which affects the performance of learning methods that rely on such data. To investigate the impact of noisy historical data on our PRBTD framework, we introduce noise to a certain proportion of the data while keeping the remaining data unchanged to create historical data with varying proportions of ground truth data. Specifically, we multiply the values of the selected data by $1+\xi$, where $\xi \sim N(0,0.1^2)$. We construct sets of historical data with 50\%, 60\%, 70\%, 80\%, and 90\% ground truth data and evaluate the performance of the PRBTD, DTI, and CNB methods on these sets. We did not include the other two baselines, TD and WEI, because they do not utilize historical data. As shown in Fig. \ref{fig-proportion}, although some biased results exist due to experimental randomness, the overall trend indicates that the effectiveness of all methods tends to decline as the proportion of ground truth in the historical data decreases. This decline is reasonable, as an increase in noise within the training data is likely to affect the prediction results of the methods. However, PRBTD consistently outperforms CNB and DTI. In practice, historical data typically undergo some level of data cleaning and enhancement, meaning that noise is usually limited. Even with 50\% of the historical data containing noise, PRBTD still performs effectively and better than traditional prediction-based methods do, demonstrating its independence from the ground truth.

\begin{table}[!t]
	\caption{Performances of the PRBTD and Baseline Methods in the Scenario with Bursty Data Values}
	\label{table-bursty}
	\centering
	\setlength{\tabcolsep}{3pt}
	\renewcommand{\arraystretch}{1.3}
	\begin{tabular}{p{70pt}|p{45pt}|p{45pt}|p{58pt}}
		\hline
		Method& F1-score & Reputation distance & Noise reduction ratio\\
		\hline
		WEI & 0.9667 & 0.0927 & 0.2657\\
		CNB & 0.9704 & 0.1309 & 0.2679\\
		TD & 0.9778 & 0.2953 & 0.4240\\
		DTI & 0.9852 & 0.1554 & 0.3726\\
		\textbf{PRBTD (ours)} & \textbf{0.9889} & \textbf{0.3686} & \textbf{0.8089}\\
		\hline
	\end{tabular}
\end{table}

\subsubsection{Experiments in Scenarios with Bursty Data Values}
In some cases, the sensing data of tasks may encompass data with bursty values, which can lead to inaccurate ground-truth predictions by prediction modules. To analyze the performance of PRBTD in scenarios with bursty data values, we randomly select 3 sets of data and reduce their values by 50\% for comparison experiments. Specifically, each set consists of data derived from 7 continuous time slots in 4 consecutive task regions. The simulation results are shown in Table \ref{table-bursty}. Furthermore, we compare the results obtained in scenarios with and without bursty data values. The comparisons presented in Fig. \ref{fig-bursty} show that in the scenario with bursty data values, the performances of the PRBTD, DTI, and CNB methods decreases, whereas those of the TD and WEI methods remain stable. These results are reasonable because the PRBTD and CNB methods rely on predictions of sensing data, whose accuracy is impacted by the occurrence of bursty data values, and DTI has a limited ability to handle sudden changes in data values. However, the performance decline of PRBTD is smaller than that of CNB, confirming that our method effectively mitigates the negative impact of prediction inaccuracies. Overall, our PRBTD framework still outperforms the other methods in terms of all three metrics, thus demonstrating strong robustness.

\begin{figure*}[!t]
	\par\hspace{6.5pt}\centering
	\begin{minipage}[b]{0.31\linewidth}
		\includegraphics[width=\linewidth]{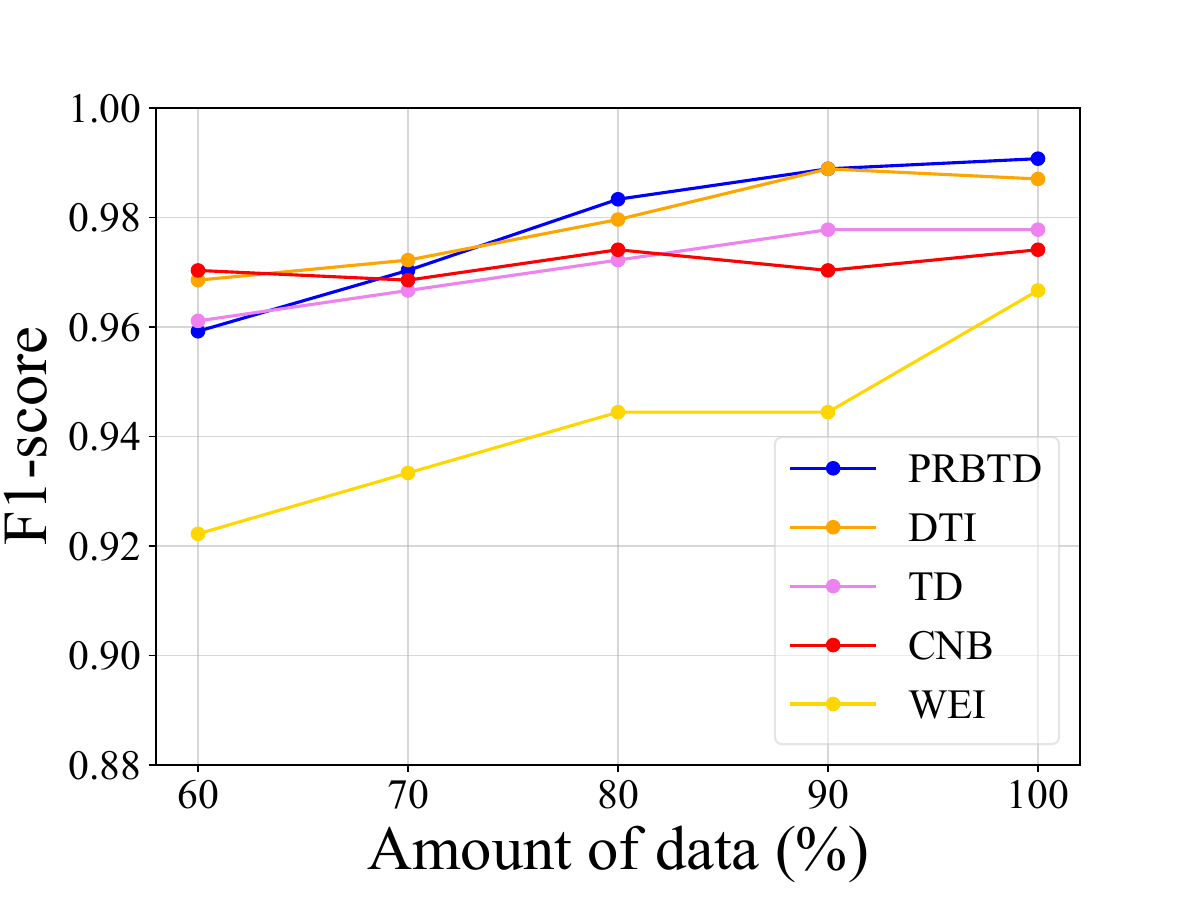}
		\centering {\fontsize{8pt}{\baselineskip}\selectfont (a) F1-score.}
		\label{fig-sparse-a}
	\end{minipage}
	\hspace{3.0mm}
	\begin{minipage}[b]{0.31\linewidth}
		\includegraphics[width=\linewidth]{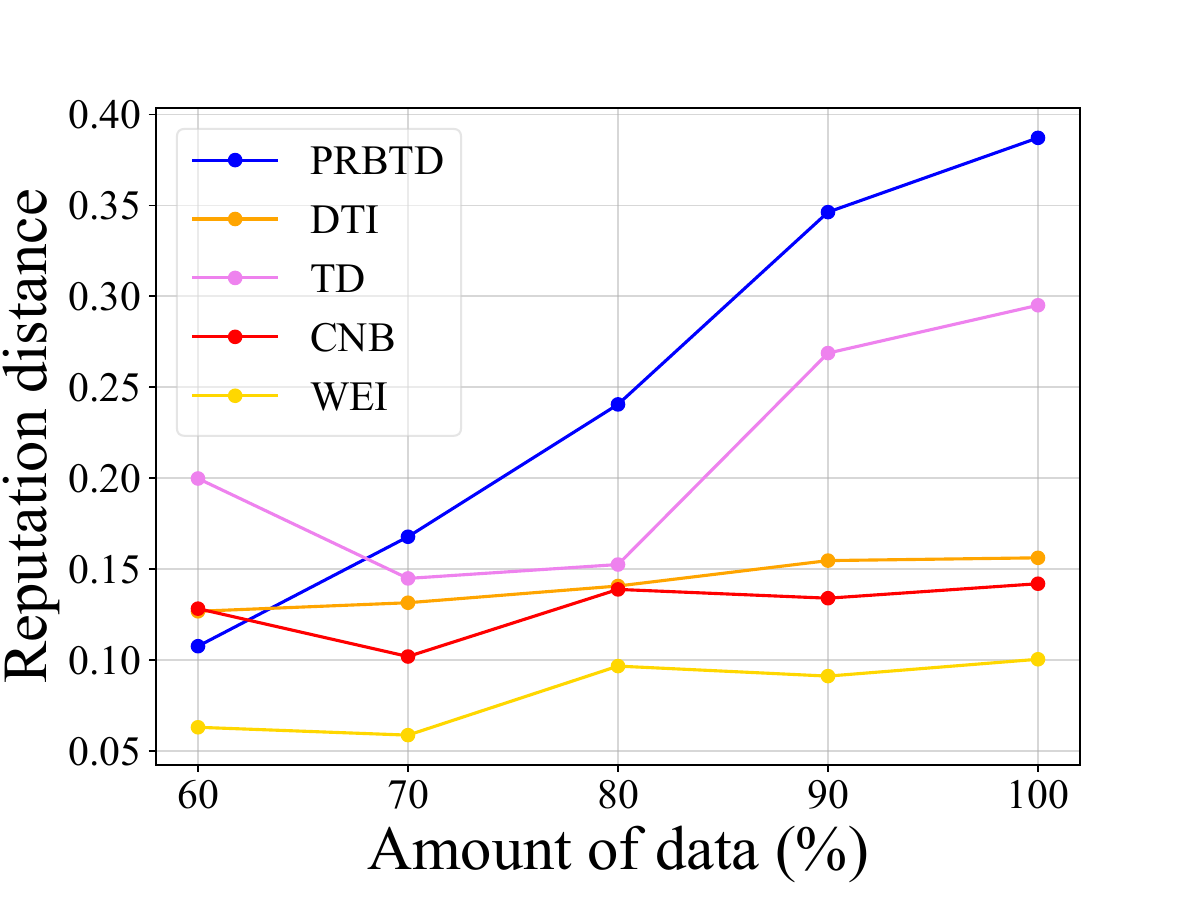}
		\centering {\fontsize{8pt}{\baselineskip}\selectfont (b) Reputation distance.}
		\label{fig-sparse-b}
	\end{minipage}
	\hspace{3.0mm}
	\begin{minipage}[b]{0.31\linewidth}
		\includegraphics[width=\linewidth]{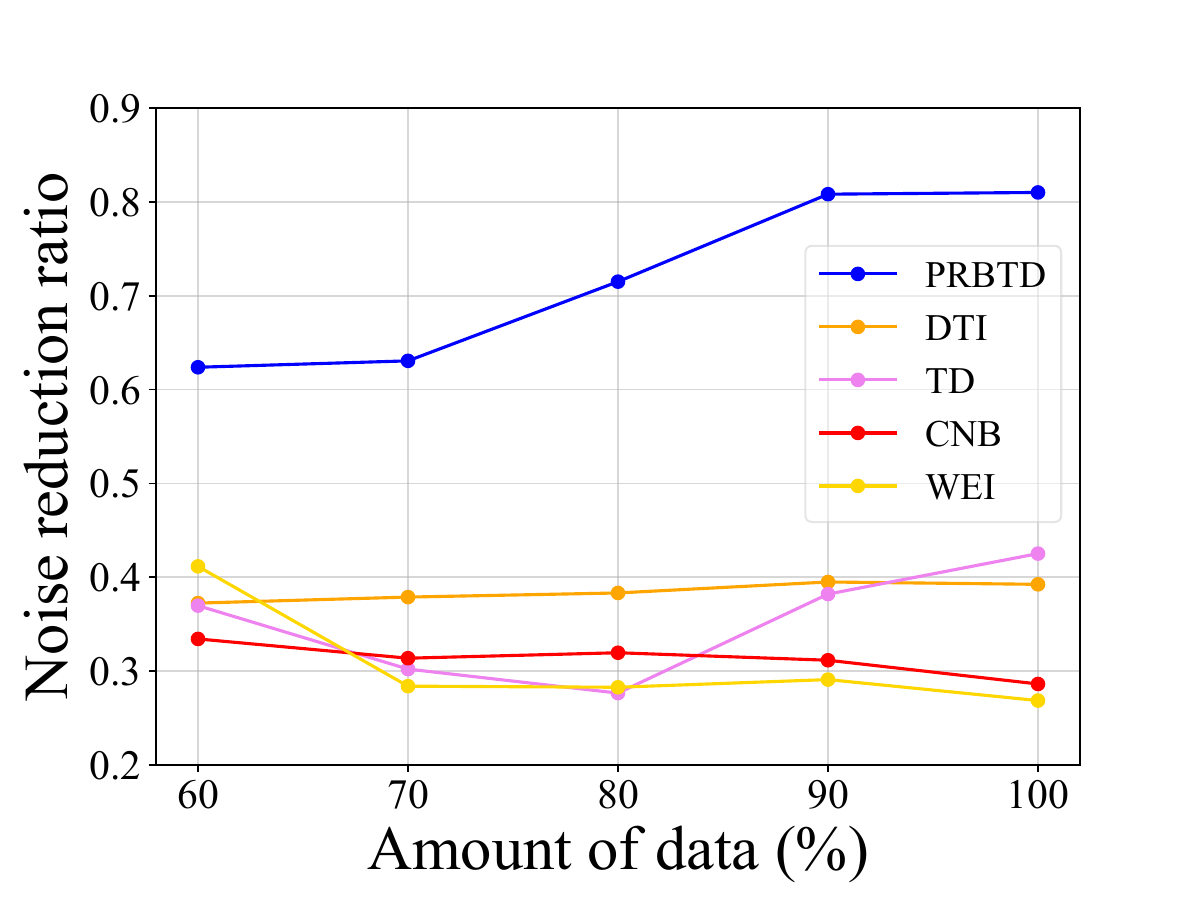}
		\centering {\fontsize{8pt}{\baselineskip}\selectfont (c) Noise reduction ratio.}
		\label{fig-sparse-c}
	\end{minipage}
	\caption{Comparisons among the PRBTD and baseline methods under different data sparsity levels.}
	\label{fig-sparse}
\end{figure*}

\begin{figure*}[!t]
	\centering
	\begin{minipage}[b]{0.30\linewidth}
		\includegraphics[width=\linewidth]{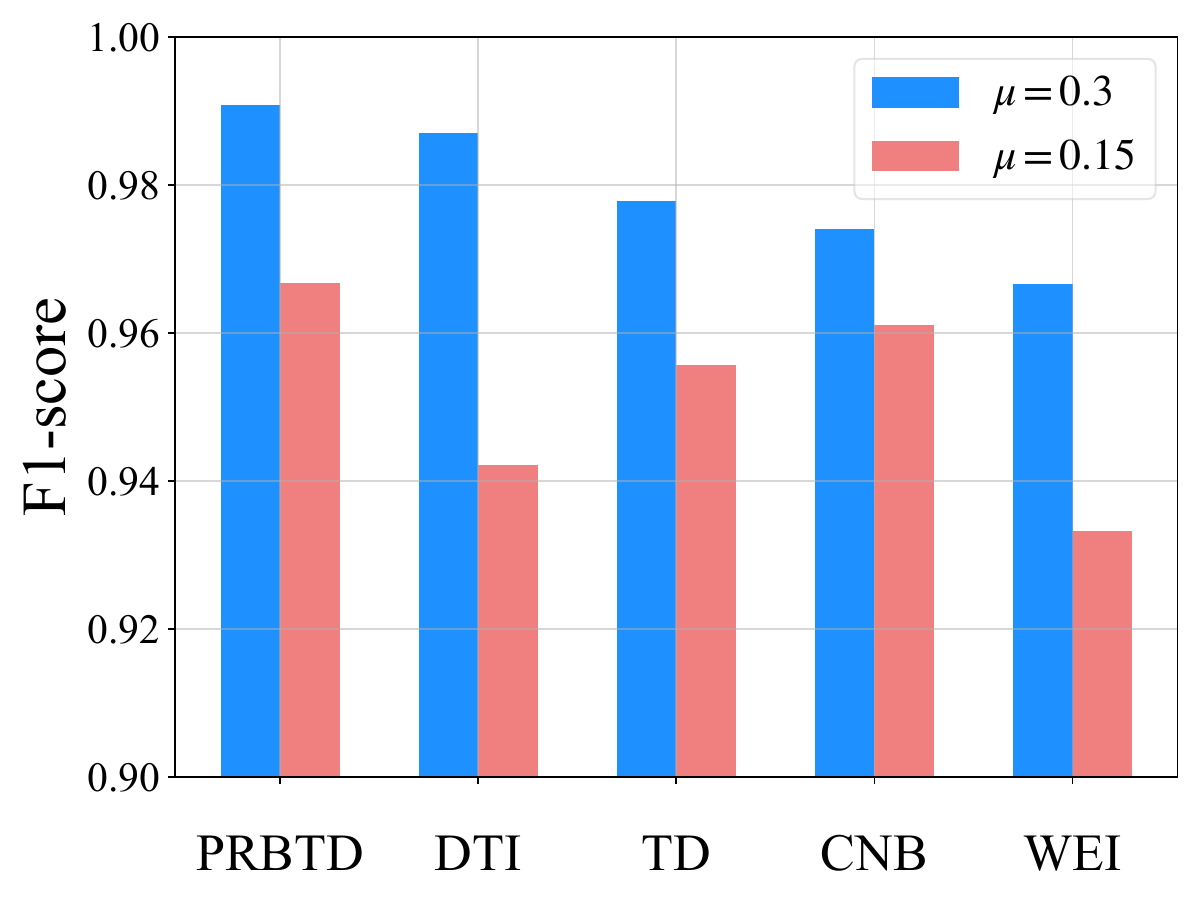}
		\par\hspace{14pt}\centering {\fontsize{8pt}{\baselineskip}\selectfont (a) F1-score.}
		\label{fig-noise-a}
	\end{minipage}
	\hspace{5mm}
	\begin{minipage}[b]{0.30\linewidth}
		\includegraphics[width=\linewidth]{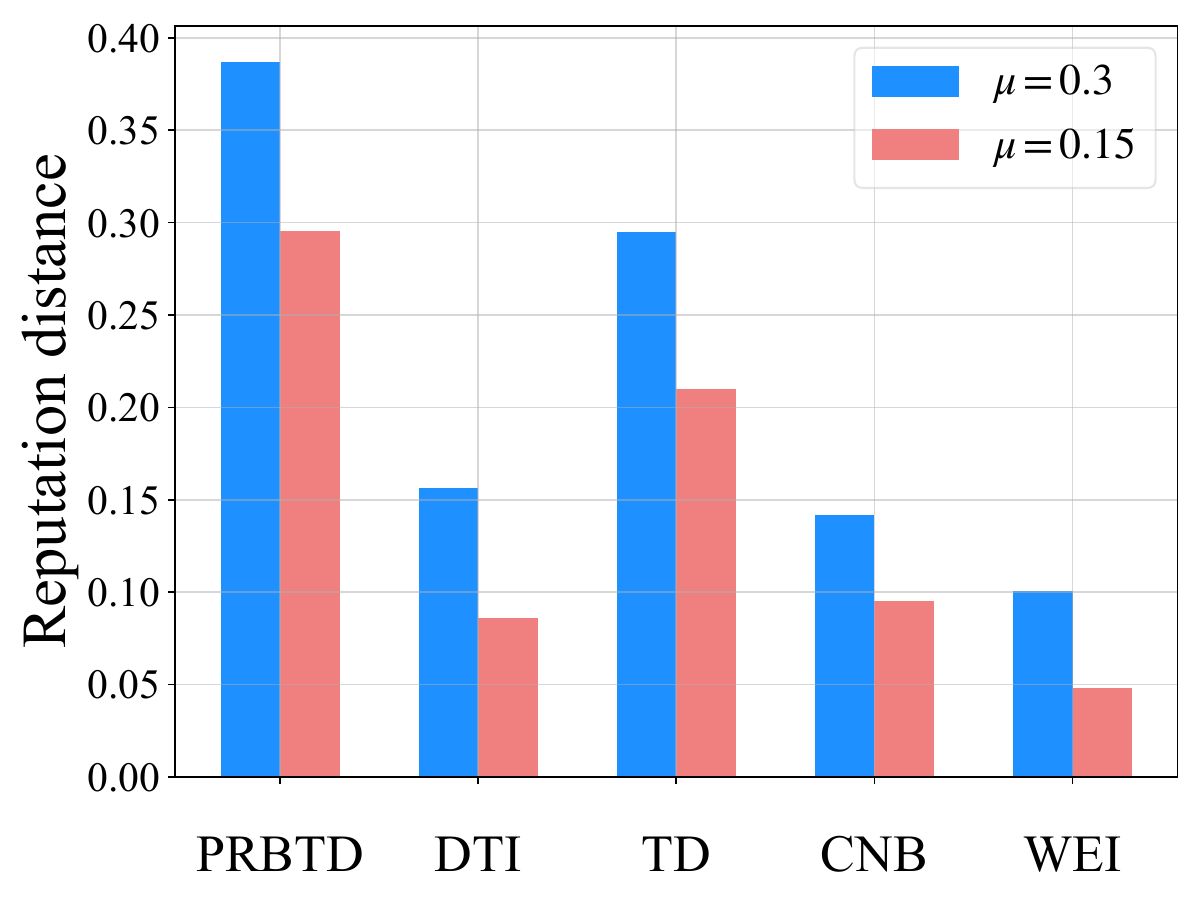}
		\par\hspace{14pt}\centering {\fontsize{8pt}{\baselineskip}\selectfont (b) Reputation distance.}
		\label{fig-noise-b}
	\end{minipage}
	\hspace{5mm}
	\begin{minipage}[b]{0.30\linewidth}
		\includegraphics[width=\linewidth]{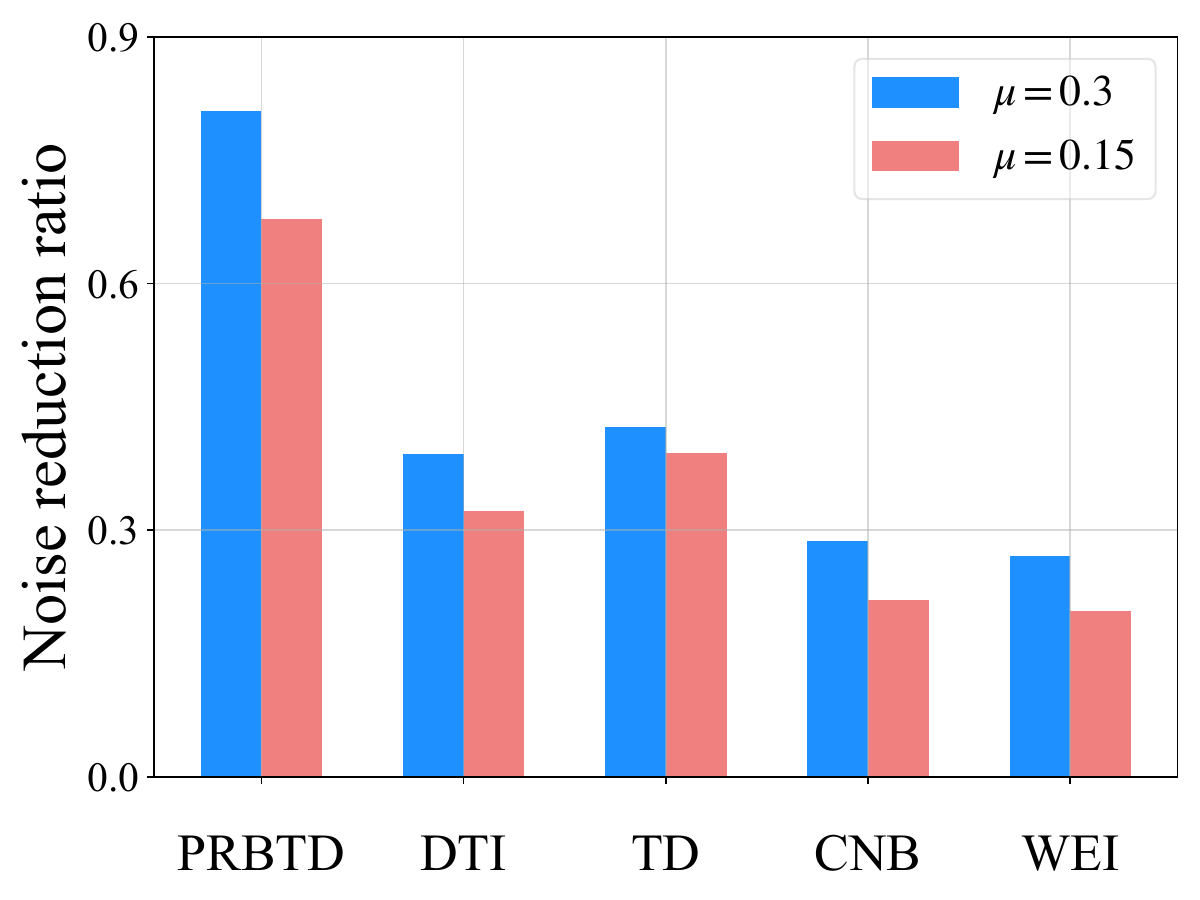}
		\par\hspace{12pt}\centering {\fontsize{8pt}{\baselineskip}\selectfont (c) Noise reduction ratio.}
		\label{fig-noise-c}
	\end{minipage}
	\caption{Comparisons among the PRBTD and baseline methods under different levels of sensing noise.}
	\label{fig-noise}
\end{figure*}

\subsubsection{Experiments in Scenarios with Sparse Data}
In MCS, the sensing data submitted by MUs may be sparse due to the uncertainty of the MU locations and period distributions, resulting in inaccurate quality evaluations in cases with insufficient data. Thus, we conduct experiments under different degrees of data sparsity by adjusting the amount of sensing data submitted by the MUs. Specifically, we analyze the performances of different methods in 5 scenarios, which use different percentages (i.e., 60\%, 70\%, 80\%, 90\%, and 100\%) of the total amount of submitted sensing data for quality evaluation purposes. The simulation results are shown in Fig. \ref{fig-sparse}. The results show that as the sparsity of the data increases, the performance of the PRBTD, DTI, TD, and WEI methods gradually decreases, whereas that of the CNB method remains stable. The reason for these declines is the reduction in the amount of available data, which leads to a decrease in the accuracy of the ground-truth estimates yielded by the PRBTD, DTI, TD, and WEI methods. Nevertheless, in most case, PRBTD still outperforms these baseline methods, as it utilizes the implications between the sensing data derived from different locations and periods, which indirectly increases the amount of data available for quality evaluation.

\begin{table}[!t]
	\caption{Performances of the PRBTD and Baseline Methods in the Scenario with Lower Levels of Noise}
	\label{table-noise}
	\centering
	\setlength{\tabcolsep}{3pt}
	\renewcommand{\arraystretch}{1.3}
	\begin{tabular}{p{70pt}|p{45pt}|p{45pt}|p{58pt}}
		\hline
		Method& F1-score & Reputation distance & Noise reduction ratio\\
		\hline
		WEI & 0.9333 & 0.0480 & 0.2022\\
		CNB & 0.9611 & 0.0950 & 0.2145\\
		TD & 0.9556 & 0.2099 & 0.3937\\
		DTI & 0.9422 & 0.0858 & 0.3229\\
		\textbf{PRBTD (ours)} & \textbf{0.9667} & \textbf{0.2954} & \textbf{0.6787}\\
		\hline
	\end{tabular}
\end{table}

\subsubsection{Experiments in Scenarios with Lower Levels of Sensing Noise}
Finally, to evaluate the effectiveness of the PRBTD method in extreme MCS scenarios, we analyze the performances of different methods in cases where the malicious MUs are more sophisticated (i.e., the sensing data submitted by the malicious MUs have lower levels of sensing noise, and some malicious MUs even submit data without noise in some time slots). Specifically, we adjust $\mu$ to 0.15 and keep the other parameters unchanged to reduce the sensing noise of the data submitted by malicious MUs. Table \ref{table-noise} and Fig. \ref{fig-noise} present and compare the performances of the different methods. The results show that with the reduction in the amount of sensing noise contained in the data derived from malicious MUs, the performances of all methods decrease because of the difficulty in identifying incorrect or fake data with less noise. However, PRBTD remains the best approach, outperforming the other methods in terms of the three metrics, thus indicating its robustness in cases involving sophisticated malicious MUs.

\section{Conclusion}
In this paper, we combine prediction, TD, and reputation mechanisms for the first time to propose a novel framework, PRBTD, for evaluating and enhancing the quality of spatio-temporally correlated sensing data in MCS systems. We propose using the implications between data points to assist in data quality evaluation and evaluate the two significant benefits of integrating implications: reducing the impact of inaccurate prediction results on data quality evaluation and increasing the amount of data available for inferring and verifying the authenticity of data. This approach makes PRBTD superior to existing methods regardless of the presence of bursty values or data sparsity.

However, our PRBTD algorithm has limitations. As its calculations rely on historical data, it cannot achieve optimal performance initially in the absence of historical data or prior knowledge. Additionally, PRBTD must learn and extract the implications between data, so it can handle data with implicit correlations only. To address these limitations, future work will focus on developing this method further to extend its applicability.

Overall, the proposed PRBTD framework makes significant advancements in the quality evaluation of sensing data in MCS systems, thereby promising more accurate and reliable data classification even in challenging scenarios.


\begin{thebibliography}{00}

\bibitem{survey}
A.~Capponi, C.~Fiandrino, B.~Kantarci, L.~Foschini, D.~Kliazovich, and
P.~Bouvry, ``A survey on mobile crowdsensing systems: Challenges, solutions,
and opportunities,'' \emph{IEEE Commun. Surveys Tuts.}, vol.~21, no.~3, pp.
2419--2465, 3rd Quart. 2019.

\bibitem{parknet}
S.~Mathur, S.~Kaul, M.~Gruteser, and W.~Trappe, ``ParkNet: a mobile sensor
network for harvesting real time vehicular parking information,'' in
\emph{Proc. ACM MobiHoc S3 Workshop}, 2009, pp.
25--28.

\bibitem{nextdoor}
Nextdoor, [Online]. Available: \url{https://nextdoor.com/}

\bibitem{googlecrowdsource}
Google crowdsource, [Online]. Available: \url{https://crowdsource.google.com/about/}

\bibitem{pp-assign}
M.~Li, J.~Wu, W.~Wang, and J.~Zhang, ``Toward privacy-preserving task
assignment for fully distributed spatial crowdsourcing,'' \emph{IEEE Internet
	Things J.}, vol.~8, no.~18, pp. 13991--14002, Sep. 2021.

\bibitem{ppta}
Z.~Wang, J.~Hu, R.~Lv, J.~Wei, Q.~Wang, D.~Yang, and H.~Qi, ``Personalized
privacy-preserving task allocation for mobile crowdsensing,'' \emph{IEEE
	Trans. Mobile Comput.}, vol.~18, no.~6, pp. 1330--1341, Jun. 2019.

\bibitem{dqnmcs}
L.~Xiao, Y.~Li, G.~Han, H.~Dai, and H.~V. Poor, ``A secure mobile crowdsensing
game with deep reinforcement learning,'' \emph{IEEE Trans. Inf. Forensics
	Secur.}, vol.~13, no.~1, pp. 35--47, Jan. 2018.

\bibitem{rst}
C.~Meng, H.~Xiao, L.~Su, and Y.~Cheng, ``Tackling the redundancy and sparsity
in crowd sensing applications,'' in \emph{Proc. 14th ACM Conf. Embedded Netw. Sens. Syst.}, 2016, pp. 150--163.

\bibitem{decentralized}
X.~Yang, Z.~Zeng, A.~Liu, N.~N. Xiong, T.~Wang, and S.~Zhang, ``A decentralized
trust inference approach with intelligence to improve data collection quality
for mobile crowd sensing,'' \emph{Inf. Sci.}, vol. 644, pp.
119286, Oct. 2023.

\bibitem{estimateQ}
S.~B. Azmy, N.~Zorba, and H.~S. Hassanein, ``Quality estimation for scarce
scenarios within mobile crowdsensing systems,'' \emph{IEEE Internet Things
	J.}, vol.~7, no.~11, pp. 10955--10968, Nov. 2020.

\bibitem{onlineQ}
X.~Zhang and X.~Gong, ``Online data quality learning for quality-aware
crowdsensing,'' in \emph{Proc. 16th Annu. IEEE Int. Conf. Sens. Commun.
	Netw.}, 2019, pp. 1--9.

\bibitem{MAB}
X.~Gao, S.~Chen, and G.~Chen, ``{MAB}-based reinforced worker selection framework
for budgeted spatial crowdsensing,'' \emph{IEEE Trans. Knowl. Data Eng.},
vol.~34, no.~3, pp. 1303--1316, Mar. 2022.

\bibitem{sttn}
Y.~Xie, J.~Niu, Y.~Zhang, and F.~Ren, ``Multisize patched spatial-temporal
transformer network for short- and long-term crowd flow prediction,''
\emph{IEEE Trans. Intell. Transp. Syst.}, vol.~23, no.~11, pp.
21548--21568, Nov. 2022.

\bibitem{stresnet}
J.~Zhang, Y.~Zheng, and D.~Qi, ``Deep spatio-temporal residual networks for
citywide crowd flows prediction,'' in \emph{Proc. 31st AAAI Conf. Artif.
	Intell.}, 2017, pp. 1655--1661.

\bibitem{starima}
X.~Min, J.~Hu, Q.~Chen, T.~Zhang, and Y.~Zhang, ``Short-term traffic flow
forecasting of urban network based on dynamic {STARIMA} model,'' in \emph{Proc.
	12th IEEE Int. Conf. Intell. Transp. Syst.}, 2009, pp. 1--6.

\bibitem{MVSTGN}
Y.~Yao, B.~Gu, Z.~Su, and M.~Guizani, ``{MVSTGN}: A multi-view spatial-temporal
graph network for cellular traffic prediction,'' \emph{IEEE Trans. Mobile
	Comput.}, vol.~22, no.~5, pp. 2837--2849, May 2023.

\bibitem{glsttn}
B.~Gu, J.~Zhan, S.~Gong, W.~Liu, Z.~Su, and M.~Guizani, ``A spatial-temporal
transformer network for city-level cellular traffic analysis and
prediction,'' \emph{IEEE Trans. Wireless Commun.}, vol.~22, no.~12, pp.
9412--9423, Dec. 2023.

\bibitem{series}
F.~Xu, Y.~Lin, J.~Huang, D.~Wu, H.~Shi, J.~Song, and Y.~Li, ``Big data driven
mobile traffic understanding and forecasting: A time series approach,''
\emph{IEEE Trans. Services Comput.}, vol.~9, no.~5, pp. 796--805, Sep./Oct. 2016.

\bibitem{kalman}
Y.~Wang, M.~Papageorgiou, and A.~Messmer, ``Real-time freeway traffic state
estimation based on extended {Kalman} filter: A case study,''
\emph{Transp. Sci.}, vol.~41, no.~2, pp. 167--181,
May 2007.

\bibitem{mcspredict}
Z.~Zhou, H.~Liao, B.~Gu, K.~M.~S. Huq, S.~Mumtaz, and J.~Rodriguez, ``Robust
mobile crowd sensing: When deep learning meets edge computing,'' \emph{IEEE
	Netw.}, vol.~32, no.~4, pp. 54--60, Jul./Aug. 2018.

\bibitem{lowmalicious}
N.~Owoh, J.~Riley, M.~Ashawa, S.~Hosseinzadeh, A.~Philip, and J.~Osamor, ``An
adaptive temporal convolutional network autoencoder for malicious data
detection in mobile crowd sensing,'' \emph{Sensors}, vol.~24, no.~7, Apr. 2024, Art. no. 2353.

\bibitem{encrytd}
Y.~Zheng, H.~Duan, and C.~Wang, ``Learning the truth privately and confidently:
Encrypted confidence-aware truth discovery in mobile crowdsensing,''
\emph{IEEE Trans. Inf. Forensics Security.}, vol.~13, no.~10, pp. 2475--2489,
Oct. 2018.

\bibitem{lighttd}
C.~Miao, L.~Su, W.~Jiang, Y.~Li, and M.~Tian, ``A lightweight
privacy-preserving truth discovery framework for mobile crowd sensing
systems,'' in \emph{Proc. IEEE Int. Conf. Comput. Commun.}, 2017, pp.
1--9.

\bibitem{sctd}
W.~Mo, Z.~Li, Z.~Zeng, N.~N. Xiong, S.~Zhang, and A.~Liu, ``{SCTD}: A
spatiotemporal correlation truth discovery scheme for security management of
data platform,'' \emph{Future Gener. Comput. Syst.}, vol. 139, pp.
109--125, Feb. 2023.

\bibitem{mcstd}
R.~W. Ouyang, M.~Srivastava, A.~Toniolo, and T.~J. Norman, ``Truth discovery in
crowdsourced detection of spatial events,'' in \emph{Proc. 23rd ACM Int.
	Conf. Inform. Knowl. Manage.}, 2014, pp. 461--470.

\bibitem{pptd}
Y.~Cheng, J.~Ma, Z.~Liu, Z.~Li, Y.~Wu, C.~Dong, and R.~Li, ``A
privacy-preserving and reputation-based truth discovery framework in mobile
crowdsensing,'' \emph{IEEE Trans. Dependable Secure Comput.}, vol.~20, no.~6,
pp. 5293--5311, Nov./Dec. 2023.

\bibitem{tf}
X.~Yin, J.~Han, and P.~S. Yu, ``Truth discovery with multiple conflicting
information providers on the web,'' in \emph{Proc. 13th ACM SIGKDD Int. Conf.
	Knowl. Discovery Data Mining}, 2007, pp. 1048--1052.

\bibitem{generaltf}
Y.~Du, H.~Xu, Y.-E. Sun, and L.~Huang, ``A general fine-grained truth discovery
approach for crowdsourced data aggregation,'' in \emph{Proc. Int. Conf. Database Syst. Adv. Appl.}, 2017, pp. 3--18.

\bibitem{epptd}
G.~Xu, H.~Li, S.~Liu, M.~Wen, and R.~Lu, ``Efficient and privacy-preserving
truth discovery in mobile crowd sensing systems,'' \emph{IEEE Trans. Veh.
	Technol.}, vol.~68, no.~4, pp. 3854--3865, April. 2019.

\bibitem{dtd}
Y.~Kang, A.~Liu, N.~N. Xiong, S.~Zhang, T.~Wang, and M.~Dong, ``{DTD}: An
intelligent data and bid dual truth discovery scheme for {MCS} in {IIoT},''
\emph{IEEE Internet Things J.}, vol.~11, no.~2, pp. 2507--2519, Jan. 2024.

\bibitem{tffds}
H.~Tian, W.~Sheng, H.~Shen, and C.~Wang, ``Truth finding by reliability
estimation on inconsistent entities for heterogeneous data sets,''
\emph{Knowl. Based Syst.}, vol. 187, Jan. 2020, Art. no. 104828.

\bibitem{design_rep}
K.~Zhao, S.~Tang, B.~Zhao, and Y.~Wu, ``Dynamic and privacy-preserving
reputation management for blockchain-based mobile crowdsensing,'' \emph{IEEE
	Access}, vol.~7, pp. 74694--74710, 2019.

\bibitem{reputation}
T.~Salman, R.~Jain, and L.~Gupta, ``A reputation management framework for
knowledge-based and probabilistic blockchains,'' \emph{Proc. IEEE Int.
	Conf. Blockchain}, 2019, pp. 520--527.

\bibitem{bcreput}
D.~C. Nguyen, M.~Ding, Q.-V. Pham, P.~N. Pathirana, L.~B. Le, A.~Seneviratne,
J.~Li, D.~Niyato, and H.~V. Poor, ``Federated learning meets blockchain in
edge computing: Opportunities and challenges,'' \emph{IEEE Internet Things
	J.}, vol.~8, no.~16, pp. 12806--12825, Aug. 2021.

\bibitem{corre}
L.~Wang, Z.~Yu, D.~Zhang, B.~Guo, and C.~H. Liu, ``Heterogeneous multi-task
assignment in mobile crowdsensing using spatiotemporal correlation,''
\emph{IEEE Trans. Mobile Comput.}, vol.~18, no.~1, pp. 84--97, Jan. 2019.

\bibitem{optim}
Y.~Wang, Z.~Cai, Z.-H. Zhan, Y.-J. Gong, and X.~Tong, ``An optimization and
auction-based incentive mechanism to maximize social welfare for mobile
crowdsourcing,'' \emph{IEEE Trans. Computat. Social Syst.}, vol.~6, no.~3,
pp. 414--429, Jun. 2019.

\bibitem{qoi}
H.~Jin, L.~Su, D.~Chen, K.~Nahrstedt, and J.~Xu, ``Quality of information aware
incentive mechanisms for mobile crowd sensing systems,'' in \emph{Proc. 16th
	ACM Int. Symp. Mobile Ad Hoc Netw. Comput.}, 2015, pp. 167--176.

\bibitem{cmabb}
M.~Xiao, B.~An, J.~Wang, G.~Gao, S.~Zhang, and J.~Wu, ``{CMAB}-based reverse
auction for unknown worker recruitment in mobile crowdsensing,'' \emph{IEEE
	Trans. Mobile Comput.}, vol.~21, no.~10, pp. 3502--3518, Oct. 2022.

\bibitem{macnb}
B.~Gu, X.~Yang, Z.~Lin, W.~Hu, M.~Alazab, and R.~Kharel, ``Multiagent
actor-critic network-based incentive mechanism for mobile crowdsensing in
industrial systems,'' \emph{IEEE Trans Ind. Informat.}, vol.~17, no.~9, pp.
6182--6191, Sep. 2021.

\bibitem{olqa}
H.~Gao, C.~H. Liu, J.~Tang, D.~Yang, P.~Hui, and W.~Wang, ``Online
quality-aware incentive mechanism for mobile crowd sensing with extra
bonus,'' \emph{IEEE Trans. Mobile Comput.}, vol.~18, no.~11, pp. 2589--2603,
Nov. 2019.

\bibitem{taskme}
B.~Guo, H.~Chen, Z.~Yu, W.~Nan, X.~Xie, D.~Zhang, and X.~Zhou, ``{TaskMe}: Toward
a dynamic and quality-enhanced incentive mechanism for mobile crowd
sensing,'' \emph{Int. J. Hum.-Comput. Stud.}, vol. 102, pp. 14--26, Jun. 2017.

\bibitem{pca}
D.~Zhang, J.~Yang, J.~Yang, and A.~F. Frangi, ``Two-dimensional {PCA}: A new
approach to appearance-based face representation and recognition,''
\emph{IEEE Trans. Pattern Anal. Mach. Intell.}, vol.~26, no.~1, pp.
131--137, Jan. 2004.

\end{thebibliography}
\end{document}